%% file: main.tex
\documentclass[twoside]{article}
\usepackage{graphicx}
\usepackage{xcolor}
\usepackage{amsmath} 
\usepackage{amssymb}
\usepackage{amsfonts}
\usepackage{geometry}
\usepackage{booktabs}
\usepackage{multirow}
\usepackage{subcaption}
\usepackage{wrapfig,lipsum,booktabs}
\usepackage{hyperref}

\hypersetup{
    colorlinks,
    citecolor=black,
    filecolor=black,
    linkcolor=black,
    urlcolor=black,
}
\usepackage[nameinlink]{cleveref}

\newcommand{\shrink}[1]{}

\usepackage[accepted]{aistats2026}

\usepackage[round]{natbib}

\begin{document}

\setlength{\abovedisplayskip}{2pt plus 3pt}
\setlength{\belowdisplayskip}{2pt plus 3pt}

\runningtitle{VIPaint: Image Inpainting with Pre-Trained Diffusion Models via Variational Inference}

\twocolumn[

\aistatstitle{VIPaint: Image Inpainting with Pre-Trained\\Diffusion Models via Variational Inference}

\aistatsauthor{ Sakshi Agarwal \And Gabriel Hope \And Jimin Heo \And Erik B. Sudderth }

\aistatsaddress{ Accenture \And Swarthmore College \And Univ.~California, Irvine \And Univ.~California, Irvine } ]

\begin{abstract}
Diffusion probabilistic models learn to remove noise added during training, generating novel data (e.g., images) from Gaussian noise through sequential denoising. However, conditioning the generative process on corrupted or masked images is challenging. While various methods have been proposed for inpainting masked images with diffusion priors, they often fail to produce samples from the true conditional distribution, especially for large masked regions. Many baselines also cannot be applied to latent diffusion models which generate high-quality images with much lower computational cost. We propose a hierarchical variational inference algorithm that optimizes a non-Gaussian Markov approximation of the true diffusion posterior. Our \textit{VIPaint} method outperforms existing approaches to inpainting, producing diverse high-quality imputations even for state-of-the-art text-conditioned latent diffusion models, 
and is also effective for other inverse problems like deblurring and superresolution.
\end{abstract}

\section{INTRODUCTION}

\input{sections/Introduction.tex}

\section{BACKGROUND}
\label{sec:background}

\vspace*{-6pt}
\subsection{Denoising Diffusion Generative Models}
\vspace*{-4pt}
\input{sections/Background.tex}

\input{sections/Inference_Background.tex}

\section{VARIATIONAL INFERENCE OF LATENT DIFFUSION PATHS}
\label{sec:VIPaint}
\input{sections/VIPaint}

\section{EXPERIMENTS \& RESULTS}
\label{sec:results}

\input{sections/Experiments_Results}

\vspace*{-2pt}
\section{CONCLUSION}
\vspace*{-2pt}
\input{sections/Conclusion}

\subsubsection*{Acknowledgements}
This research was supported in part by NSF Robust Intelligence Award No.~IIS-1816365, ONR Award
No.~N00014-23-1-2712, and the HPI Research
Center in Machine Learning and Data Science at UC Irvine.

\newpage
\bibliographystyle{apalike} 
\bibliography{aistats}

\clearpage
\appendix
\onecolumn
\input{sections/appendix}

\end{document}

%% file: sections/Introduction.tex
Diffusion models ~\citep{hoddpm, song2021scorebased, ddpm++, smld}  learn to generate synthetic data by sequentially reducing Gaussian noise across hundreds or thousands of steps, producing deep generative models that have advanced the state-of-the-art in natural image generation~\citep{dhariwal2021diffusion,kingma2021variational,edm}. Diffusion models for high-dimensional data like images are computationally intensive. Efficiency may be improved by leveraging an autoencoder~\citep{kingma2019introduction, Rombach_2022_CVPR, vahdat2021score} to map data to a lower-dimensional space, and then training a diffusion model for the lower-dimensional codes.  Such \emph{latent diffusion models} (LDMs) enable efficient but expressive models for images with millions of pixels.

\begin{figure}[t] %
    \centering
    \small 
\includegraphics[width=.99\linewidth]{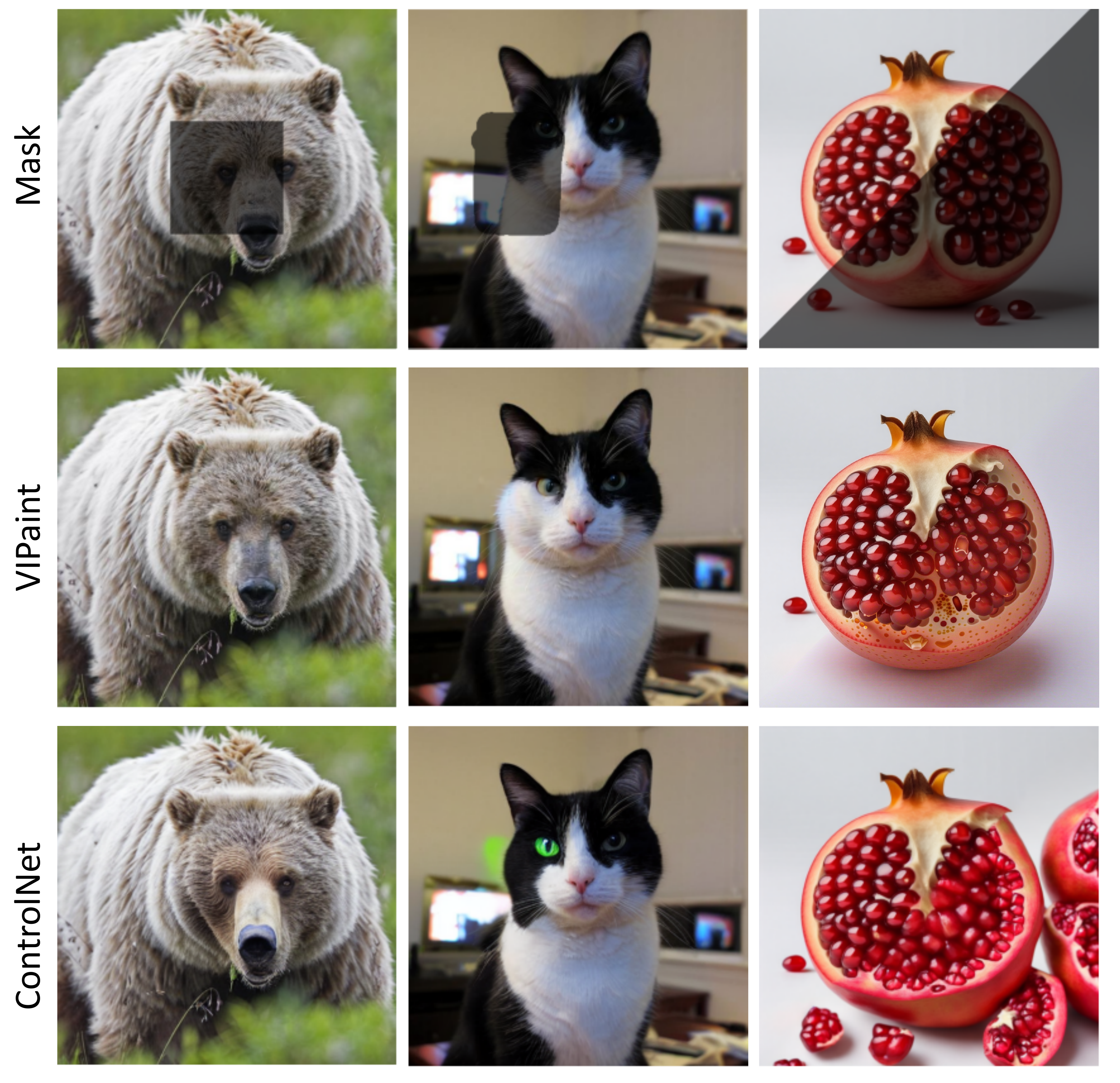}
\vspace*{-18pt}
    \caption{Image inpainting by applying our VIPaint method to pre-trained Stable Diffusion 3.5, and a model retrained via ControlNet \citep{zhang2023adding}.}
\label{fig:stable_diffusion}
\end{figure}

Motivated by the foundational information captured by diffusion models, numerous algorithms have used a pre-trained diffusion model as a prior for tasks such as image editing~\citep{meng2021sdedit}, inpainting~\citep{song2021scorebased,wang2023zeroshot,ddrm,chung2022improving,Lugmayr2022RePaintIU,cardoso2024monte, Feng_2023_ICCV,trippe2023diffusion,dou2024diffusion}, and other inverse problems~\citep{NEURIPS2021_6e289439,song2023pseudoinverseguided,graikos2022diffusion,mardani2024variational,chung2023diffusion}.
Many of these methods are specialized to inpainting with pixel-based diffusion models, a simpler task where every pixel is either perfectly observed or completely missing, and are not easily adapted to state-of-the-art LDMs. Moreover, they 
are often tested on relatively \emph{easy} restoration tasks where much of the image's global structure is already known. Methods like DPS~\citep{chung2023diffusion} and REDdiff~\citep{mardani2024variational} are most effective at inpainting small masked regions, or for tasks like deblurring and super-resolution that only require the refinement of local image details. %

RePaint~\citep{Lugmayr2022RePaintIU} and CoPaint~\citep{pmlr-v202-zhang23q} observe that existing methods often produce inconsistent or unrealistic inpaintings in large, contiguous masked regions. %
Most methods employ an iterative refinement procedure, like that used to generate unconditional samples, and guide their predictions towards the partially observed image via various approximations and heuristics. We hypothesize that this sequential denoising process, from independent Gaussian noise to noise-free images, tends to misrepresent the global structure early in the reverse diffusion trajectory. As the process lacks a mechanism to correct these early errors, %
the final inpainting can remain globally incoherent when inpainting large regions.

Recent work proposing similar algorithms for image editing~\citep{Avrahami_2022_CVPR} or inpainting~\citep{rout2023solving, Corneanu_2024_WACV, chung2023diffusion, resample, DAPS} with LDMs %
suffers from similar inaccuracies (see Sec.~\ref{sec:results}). \cite{liu2024image} adapt probabilistic circuits~\citep{ProbCirc20} for large-mask image inpainting, but their supervised approach must be trained to match a known image mask distribution. Methods based on generative adversarial networks~\citep{zhao2021comodgan_comodgan,lama} do not leverage foundation models and have similar restrictions, requiring specialized training and many examples of similarly corrupted images for each task. %
\cite{wang2024magic} assume additional side-information, such as segmentations or depths or poses, is available to inpaint large mask regions.

On the other hand, \emph{variational inference} (VI)~\citep{wainwright08,blei2017variational} has achieved excellent image restoration results with a wide range of priors, including mixtures~\citep{fergus2006removing,ji2017patches} and hierarchical VAEs~\citep{pmlr-v216-agarwal23a}, but there is little work exploring its integration with state-of-the-art LDMs. While \emph{REDdiff}~\citep{mardani2024variational} applies VI to approximate the posterior of pixel-based DMs, its local approximation of the noise-free image posterior is difficult to optimize, requiring annealing heuristics that are sensitive to local optima. MGPS~\citep{moufadvariational} uses VI locally within a sequential denoising procedure to approximately sample from the posterior at a midpoint between the current noise level and the noise-free image. 

In this work, we propose \emph{VIPaint}, a novel application of VI that employs both LDMs and pixel-based DMs as priors to handle challenging inference problems, such as large mask image inpainting. VIPaint strategically defines a hierarchical, Markovian and non-Gaussian approximation to the true (L)DM posterior that accounts for a subset of latent noise levels, enabling the inference of both high-level semantics and low-level details from observed pixels \emph{simultaneously}.
We efficiently infer variational parameters for each inpainting query, avoiding the need to collect a training set of corrupted images~\citep{liu2024image, Corneanu_2024_WACV}, expensively fine-tune generative models~\citep{Avrahami_2022_CVPR} or normalizing flow posteriors~\citep{Feng_2023_ICCV} for each query, or retrain large-scale conditional diffusion models~\citep{Rombach_2022_CVPR,saharia2022palette,nichol2022glide,Chung_2022_CVPR}.  

A primary design goal for VIPaint is to avoid the training of a specialized model for each image restoration task, and to instead apply pretrained DMs in a ``zero shot'' fashion, via variational inference.  Our experimental comparisons are thus primarily to the rich literature of methods that adapt pretrained DMs in other ways, but we find that VIPaint may nevertheless outperform specialized inpainting algorithms; see Fig.~\ref{fig:stable_diffusion}.

\vspace*{-2pt}
We begin by reviewing properties of (latent) diffusion models %
and prior work on inference with pre-trained diffusion models in Sec.~\ref{sec:background}.  Sec.~\ref{sec:VIPaint} then develops the VIPaint algorithm, which first fits a hierarchical posterior that best aligns with the observations, and then samples diverse reconstructions from this posterior.  Results in Sec.~\ref{sec:results} on inpainting, as well as linear deblurring and superresolution, %
show substantial qualitative and quantitative improvements in producing inpaintings that capture the richness of contemporary DMs. %

%% file: sections/Background.tex
The diffusion process begins with clean data $x$, and defines a sequence of increasingly noisy representations of $x$. We denote these \emph{latent variables} by $z_t \in\{z_0,...,z_T\}$, where the \emph{time} $t$ runs from $t=0$ (low noise) to $t=T$ (substantial noise). The distribution of $z_t$ given $x$, for any  time $t \in [0,T]$, is  
\begin{equation}
q(z_t \mid \bar{x}=\texttt{enc}(x)) = \mathcal{N}(z_t \mid \alpha_t \bar x, \sigma_t^2 I), 
\label{eq:forward}
\end{equation}
where $\sigma_t > 0$ and strictly increases with $t$. 
For LDMs, $\bar{x}=\texttt{enc}(x)$ where $\texttt{enc}(\cdot)$ is a pre-trained encoder function that maps $x$ to a lower-dimensional latent code for sampling efficiency.  
For pixel-based DMs, $\texttt{enc}(x)=x$.

Diffusion induces a Markov chain for which the conditionals $q(z_t \mid z_{s})$, $q(z_{s} \mid z_{t}, \bar{x})$ are tractable Gaussians for any $s < t$ (see App.~\ref{app:diff_recap}). 
The signal-to-noise ratio~\citep{vdms} induced by this diffusion process at time $t$ equals $\text{SNR}(t) = \alpha_t^2/\sigma_t^2$. The SNR monotonically decreases with time, so that $\text{SNR}(t) < \text{SNR}(s)$ for $t > s$. %
This DM specification includes variance-preserving diffusions~\citep{hoddpm, sohl-dickstein} where $\alpha_t = \sqrt{1 - \sigma_t^2}$, variance-exploding diffusions~\citep{smld, song2021scorebased} where $\alpha_t = 1$, and rectified flow models~\citep{liu2022flow} where $\alpha_t = 1 - \sigma_t$.

\textbf{Image Generation.} We generate novel images by reversing the diffusion process of Eq.~\eqref{eq:forward}, inducing a hierarchical generative model that samples a sequence of latent variables $z_t$ before sampling $x$. Generation progresses backward in time from $t=T$ to $t=0$ via a finite temporal discretization into $T \approx 1000$ steps, either uniformly spaced as in discrete diffusion models~\citep{hoddpm}, or via a possibly non-uniform discretization~\citep{edm} of an underlying continuous-time stochastic differential equation~\citep{song2021scorebased}. Letting $t-1$ be the timestep preceding~$t$, %
the generative model for data $x$ is expressed as:
\begin{equation}
p_\theta(x) = \int_z p(z_T)  p(x \mid z_0)  \prod_{t=1}^T p_\theta(z_{t-1} \mid z_{t}) \;dz.
\label{eq:reverse}
\end{equation}
The marginal distribution of $z_T$ is typically a spherical Gaussian $p(z_T) = \mathcal{N}(z_T \mid 0, \sigma_T^2 I)$. Pixel-based diffusion models take $p(x \mid z_0)$ 
to be a simple factorized likelihood~\citep{kingma2021variational} for each pixel in $x$, $p(x \mid z_0)\propto q(z_0\mid x)$, while LDMs define $p(x \mid z_0)$ via a pre-trained decoder neural network so that $\mathbb{E}[x\mid z_0]=\texttt{dec}(z_0)$. The conditional latent distribution $p_\theta(z_{t-1}\mid z_t)$ maintains the Gaussian form $q(z_{t-1} \mid z_t, \bar{x})$ induced by the forward noise process, 
\begin{equation}
\begin{split}
    p_\theta(z_{t-1} \mid z_t) = q(z_{t-1} \mid z_t, \bar x = \hat x_\theta( z_t,t)) \\
   \text{where} \quad \hat x_\theta( z_t, t) = \frac{z_t - \sigma_t \hat \epsilon_\theta(z_t,t)}{\alpha_{t}},
\end{split}
\label{eq:one-step-denoising_a}
\end{equation}
but with the encoded data $\bar{x}=\texttt{enc}(x)$ approximated via a differentiable noise predictor %
$\hat \epsilon_\theta(z_t, t)$.  The denoising neural network may incorporate U-Net~\citep{unet} or transformer \citep{peebles2023scalable} architectures, %
and is trained to optimize a variational lower bound~\citep{hoddpm,song2021scorebased} of the marginal likelihood of data $x$,
\begin{align}
    -&\log p_\theta(x) 
    \leq \mathcal{L}(\theta; x) = C +
\label{eq:kl} \\
    & \frac{T}{2} \, \mathbb{E}_{t,\epsilon, x} \bigg[
       \left( \frac{\sigma_t^2 \alpha_{t-1}^2}{\alpha_t^2 \sigma_{t-1}^2} - 1 \right)
       \|\epsilon - \hat \epsilon_\theta(\alpha_t \bar x + \sigma_t \epsilon, t)\|_2^2
    \bigg], %
\nonumber
\end{align}
for a constant $C$.
The expectation is over $x \sim p_{\text{data}}(x)$, $\epsilon \sim \mathcal{N}(0, I)$, $t\sim \text{Uniform}(1, T)$.
Some DMs drop the SNR weights in~\eqref{eq:kl} during training~\citep{hoddpm}.

%% file: sections/Inference_Background.tex
\textbf{Inference with Diffusion Models.}
\label{sec:inference}
Image inpainting (or more broadly, data imputation) tasks arise when we are given partial observations $y=x\odot m$, where $m$ is a binary mask indicating missing pixels. %
Recovering $x$ from $y$ is challenging, especially when large image regions are masked, because many $x$ could produce the same observation $y$. Inpainting is an example of a broader class of \textit{linear inverse problems} which also includes tasks such as deblurring and super-resolution. To express the posterior $p_\theta(x \mid y)$ given a DM prior, we adapt the %
generative process of Eq.~\eqref{eq:reverse}:
\vspace*{-8pt}
\begin{equation}
    p_\theta(x|y) = \int_z p_\theta(z_T|y) p_\theta(x | z_0,y)  \prod_{t=1}^T p_\theta(z_{t-1}|z_{t},y )\;dz.
\label{eq:conditionalsampling}
\end{equation} 
Exactly evaluating this predictive distribution is infeasible due to the non-linear noise prediction and decoder networks, which make 
$p_\theta(z_{t-1} \mid z_t, y)$ %
intractable.  Several heuristic methods for approximately sampling from $z$ given $y$ are discussed in App.~\ref{app:diff_recap}. %
We instead develop a variational inference algorithm that more accurately approximates $p_\theta(x,z \mid y)$.

\vspace*{-5pt}
\subsection{Variational Inference of Missing Data} %
\vspace*{-5pt}

\emph{REDdiff} \citep{mardani2024variational} %
uses pixel-based DMs as priors and defines a Gaussian variational distribution 
$p_\theta(x\mid y) \approx q_\lambda(x) = \mathcal{N}(x \mid \mu, \sigma^2 I)$ over the data space, 
where $\lambda = \mu$ and 
the variance is fixed to the same small constant $\sigma^2 \approx 0$ for all pixels. By reducing the posterior approximation to a \emph{single} image $\mu$, REDdiff induces a simple variational inference objective: %
\begin{align}
    D\big(q_\lambda(x) \,\|\, p(x | y)\big) 
    &= - \log p(y | \mu) 
       + D\big(q_\lambda(x) \,\|\, p_\theta(x)\big). \label{eq:reddiff}
\end{align}
REDdiff seeks an image $\mu$ that reconstructs the observation $y$ (at the pixels not occluded by the mask $m$), while simultaneously having high probability (low KL divergence) under the prior.
This diffusion regularizer decomposes as an expectation over many times.

While the loss of Eq.~\eqref{eq:reddiff} is simple, \cite{mardani2024variational} find direct optimization to be difficult and unstable, and  find that annealing time from $t=T$ to $t=0$ (as in standard diffusion samplers) outperforms 
unbiased optimization of the variational bound through random time sampling. Visual examples in 
the Appendix %
compare REDdiff-V, which uses random-time sampling as justified by the correct variational bound, and REDdiff which gradually anneals time from $T$ to $0$. REDdiff also does not propagate gradients through the denoising network $\epsilon_\theta(z_t, t)$, as optimization of the true variational bound would require, to prevent optimization instability. We hypothesize that this instability is due to the denoising function's lack of smoothness at low noise levels \citep{yang2024lipschitz}. 

\begin{figure*}[t]
\small 
\centerline{\includegraphics[width=0.99\linewidth]{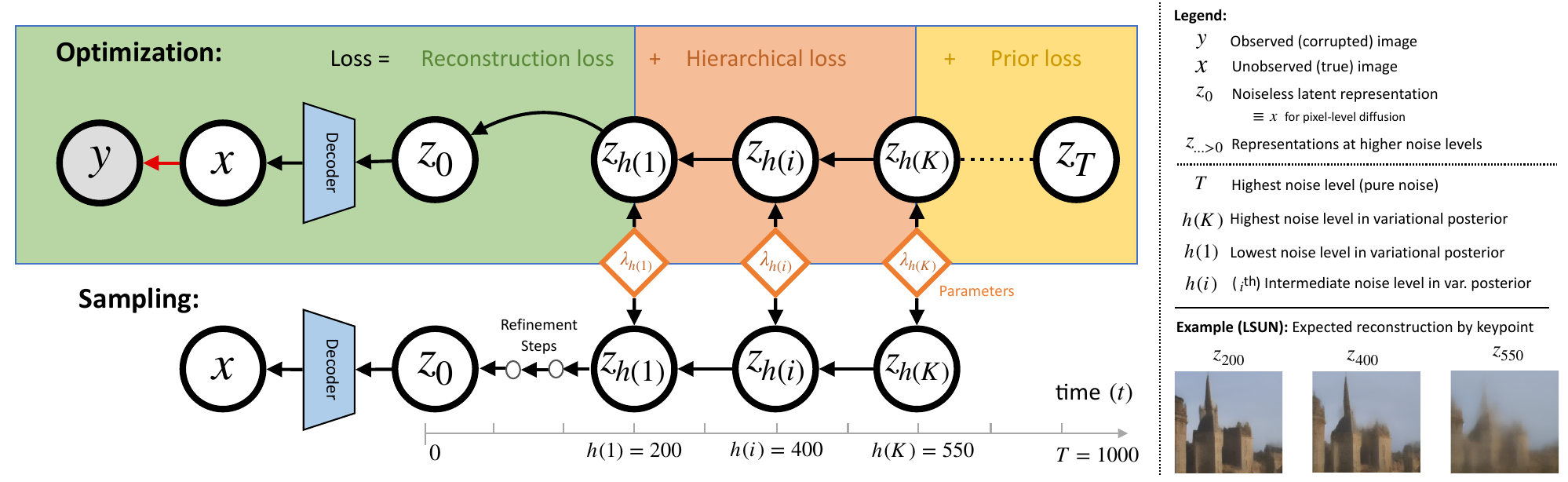}}
\vspace*{-10pt}
\caption{ The hierarchical approximate posterior of VIPaint is defined over a coarse sequence of intermediate latent steps, or keypoints,  between $h(K)$ and $h(1)$. During optimization, the variational parameters $\lambda$ defining the posterior at these sparse times are fit via a prior loss on times above $h(K)$, a hierarchical loss defined across $K$ keypoints, and a reconstruction loss estimated using a sample-based one-step approximation of $p_\theta(x \mid z_{h(1)})$. %
After a single variational optimization, multiple samples may be drawn via gradient-based stochastic refinement.
}
\label{fig:vipaint-description}
\end{figure*}

Because REDdiff employs a simple variational posterior that directly optimizes an image at the noise-free ($t=0$) level only, it is inherently incapable of capturing uncertainty in $x$, and instead seeks a single posterior mode. Additionally, its optimization process is \emph{biased} because it relies on annealing time during the diffusion process rather than randomly sampling time points. We demonstrate that our VIPaint framework better models posterior uncertainty, enables stable optimization of an unbiased variational bound, and is applicable to both pixel-based and latent DMs.

%% file: sections/VIPaint.tex
Given a pre-trained diffusion model, VIPaint approximates the distribution of latent variables $z_t$ given a test observation $y$. It defines a $K-$step hierarchical posterior over a coarse sequence of mid-ranged timesteps $\{h(i)\}_{i=1}^K$, where $h(1)$ and $h(K)$ mark the start and end timepoints in the hierarchy. This posterior is defined by variational parameters $\lambda$ optimized, for each image without amortization~\citep{kingma2019introduction}, via a variational bound. After optimization, we 
perform iterative, gradient-based refinements of posterior samples in the low-noise latent space $[0,h(1)]$ to generate  final inpaintings. See Fig.~\ref{fig:vipaint-description} for an overview. 

VIPaint offers several practical advantages over REDdiff~\citep{mardani2024variational}, as the latent-space hierarchical posterior: 1) infers global-to-local semantics in the latent space, consistent with the corrupted image $y$; 2) accounts for uncertainty in missing pixels; 3) strategically avoids training instabilities~\citep{yang2024lipschitz} which arise in the low-noise latent space $[0, h(1))$; and 4) easily extends to latent DMs. %
Below, we detail VIPaint's Markov posterior, optimization, and sampling strategies for diverse inpaintings.

\textbf{Variational Posterior Formulation.}
VIPaint defines a latent-space hierarchical posterior over a subset of $K$ \emph{keypoints} %
that capture the most informative phases of the reverse diffusion process:
\begin{equation*}
 q_\lambda(z_{h(1):h(K)})
 = \Bigg ( \prod_{i=1}^{K-1} q_{\lambda}(z_{h(i)} \mid z_{h(i+1)}) \Bigg)  q_\lambda(z_{h(K)}).
\end{equation*}
Here, $K\geq2$ and $h(i)$ is the time of the keypoint preceding $h(i+1)$. %
Experiments suggest tuning these keypoints to capture intermediate-noise timesteps with SNR %
($\alpha_t^2/\sigma_t^2$) in the range $[0.2, 0.5]$ across different (latent) DMs, see Sec~\ref{sec:hyperparameters}. For the highest timestep $h(K)$, we let 
$q_\lambda(z_{h(K)}) = \mathcal{N}(z_{h(K)} \mid \mu_{h(K)}, \tau_{h(K)})$  
be a factorized Gaussian. 
For lower-noise timesteps,  %
\begin{align}
  & q_\lambda(z_{h(i)} \mid z_{h(i+1)}) \nonumber \\
  &= \mathcal{N} (z_{h(i)} \mid \gamma_{h(i)} \bar z_{h(i)} + (1-\gamma_{h(i)}) \mu_{h(i)}, \tau_{h(i)}^2 ),
  \label{eq:VIPaint_pos} \\
  &\text{ }
  \bar z_{h(i)} = E[z_{h(i)}\mid z_{h(i+1)}, \bar x =\hat x_\theta(z_{h(i+1)}, h(i+1))]. 
  \nonumber
\end{align} 
The standard deviation $\tau_{h(i)}$ is learned and varies across dimensions of the data (or latent code for LDMs), allowing the posterior to dynamically increase uncertainty in regions with more masked image pixels.  The mean is a convex combination (with learned, dimension-specific weights $\gamma_{h(i)}$) of the prior diffusion prediction 
$\bar z_{h(i)}$
and an image-specific variational parameter $\mu_{h(i)}$.  Intuitively, optimizing $\mu_{h(i)}$ allows predictions from the DM prior to be perturbed towards means consistent with the observation $y$.  Note that the overall variational approximation is Markov (like the true posterior) but non-Gaussian, due to the incorporation of the non-linear denoising network $\hat x_\theta(\cdot)$.

Some sampling-based inpainting methods~\citep{song2021scorebased, Lugmayr2022RePaintIU, ddrm, resample} also linearly combine observations $y$ with samples $z_t$, but employ either hard constraints or manually-tuned weights. VIPaint instead incorporates free parameters $\lambda = \{ \mu_{h(K)}, \tau_{h(K)}, (\gamma_{h(i)}, \mu_{h(i)}, \tau_{h(i)})_{i=1}^{K-1} \}$ across $K$ latent levels, defined over each pixel in the image or its encoding. This flexible posterior is key to effectively reusing the diffusion prior \emph{and} aligning precisely with a particular observation $y$, without the need to re-train a specialized, conditional DM. %
We use $y$ to initialize $\mu_{h(i)}$ by scaling its encoding $\texttt{enc}(y)$ by the forward diffusion constant $\alpha_{h(i)}$.  We use %
the DM noise schedule to automatically initialize posterior variances, see App.~\ref{sec:exp-details}. 

\textbf{Fitting the Variational Posterior.}
We optimize a variational lower bound on the marginal likelihood of the observation $y$. 
As derived in App.~\ref{app:vipaint_math}, 
\begin{equation}
    \begin{split}
        L&(\lambda) = \underbrace{- \mathbb{E}[\log p_\theta(y|z_{h(1)})]}_\text{reconstruction loss}  +  \underbrace{ D\Big[q_\lambda(z_{h(K)}) || p_\theta(z_{h(K)})  \Big]}_\text{prior loss} \\
   + &\underbrace{\sum_{i=1}^{K-1} \mathbb{E}_{z_{h(i+1)}} D\Big[ q_\lambda(z_{h(i)} | z_{h(i+1)})~||~p_\theta(z_{h(i)} | z_{h(i+1)} ) \Big]}_\text{hierarchical loss}.
    \end{split}
\label{eq:VIPaintobj}
\end{equation}
VIPaint seeks latent-posterior distributions that assign high likelihood to the observed features $y$ (by minimizing the reconstruction loss), while simultaneously aligning with the medium-to-high noise levels encoding image semantics (hierarchical and prior losses). %
Expectations are with respect to the hierarchical approximate posterior $q_\lambda(z_{h(1) : h(K)})$. We approximate $L(\lambda;y )$ with $M$ Monte Carlo samples (typically 5-10) from $q_\lambda(z_{h(1) : h(K)})$, via an ancestral sampler that proceeds from high to low noise: $z_{h(K)}^{(m)} \sim q_\lambda(z_{h(K)}), z_{h(i)}^{(m)} \sim q_\lambda(z_{h(i)} | z_{h(i+1)}^{(m)})$ for $i=K-1,\ldots,1$. 
We use automatic differentiation, with reparameterized~\citep{kingma2019introduction} representation of samples from $q_\lambda$, to compute gradients with respect to $\lambda$. This allows end-to-end optimization through multiple applications of the denoising network $\hat x_\theta(\cdot)$, with no annealing.

\textbf{Reconstruction Loss.}
This term guides the posterior to align its samples $z_{h(1)}$ with observations $y$, but it is intractable to analytically integrate over both $x$ and $z_0$. Estimation via sampling is possible, but expensive as it would require backpropagation through multiple %
sampling steps. We instead adopt the approximation of \citep{rout2023solving, chung2023diffusion}: 
\begin{equation}
    p(y\mid z_{h(1)}) \approx p(y \mid \texttt{dec}(\hat x_\theta (z_{h(1)}, h(1))).
\end{equation}
For non-latent DMs, $\texttt{dec}(x)=x$. %
As $h(1)$ is close to $t=0$, this approximation leads to updates in the posterior parameters accurate enough to guide samples of $z_{h(1)}$ to be consistent with $y$. As discussed below, after optimization samples of $z_0$ are drawn conditioned on $y$ and $z_{h(1)}$ using a more fine-grained process that samples \emph{all} intermediate steps between $z_{h(1)}$ and $z_0$. 

We use the $L_1$ reconstruction loss (Laplace log-likelihood) for $\log p(y\mid x)$, and add a perceptual loss term \citep{zhang2018perceptual} when using LDMs, to better match the objective originally used to train the decoder and reduce blur (see Appendix).

\textbf{Prior Loss.} As derived in the Appendix, %
\begin{align}
    D&\Big(q_\lambda(z_{h(K)}) \,\|\, p_\theta(z_{h(K)})\Big) =
\label{eq:diff_loss}
    \\
    &\frac{T - h(K)}{2}\, \mathbb{E} \Big[ D\big(q(z_{t-1}| z_t, z_{h(K)}) \,\|\, p_\theta(z_{t-1}|z_t)\big) \Big].
\nonumber
\end{align}

The expectation is over $t$ %
uniformly sampled in $[h(K), T]$ instead of the entire range $[0, T]$, $z_{h(K)} \sim q_\lambda(z_{h(K)})$, and $z_t \sim q(z_t \mid z_{h(K)})$.
This loss regularizes the samples $z_{h(K)}$ to follow the high-level image semantics implicitly encoded by diffusions in $[h(K), T]$. 

 \textbf{Hierarchical Loss.}  The hierarchical loss term further regularizes posterior samples $\{ z_{h(i)}^{(m)}\}_{i=1}^{K-1}$ to capture high-to-mid-level image details in the critical intermediate range~\citep{edm} of noise levels $[h(1),h(K)]$ in the latent diffusion space. As with the prior loss, the hierarchical loss can be estimated via sampling (see Appendix);
 given a sample $z_{h(i+1)}^{(m)}$, $D\Big( q_\lambda(z_{h(i)} \mid z_{h(i+1)}^{(m)})~||~p_\theta(z_{h(i)} \mid z_{h(i+1)}^{(m)} ) \Big)$ can be computed analytically. %
 We simplify computation by aligning the time discretization of the prior to the posterior keypoints, %
 as in methods for accelerating unconditional DM sampling~\citep{ddim}, and thus avoid the need to sample times between keypoints. 
 
\textbf{Optimization.}
The number of optimization steps may be chosen to flexibly trade speed for accuracy.  %
If the posterior is only defined on the noise-free level $z_0$ as in REDdiff~\citep{mardani2024variational}, the VIPaint objective of Eq.~\eqref{eq:VIPaintobj} degenerates to their (non-annealed) variational bound. However, VIPaint strategically avoids low noise levels in its posterior,  avoiding the instabilities that substantially reduce REDdiff performance, and enabling generalization to LDMs.

\textbf{Sampling.}
After optimization, the hierarchical posterior $q_\lambda(z_{h(1):h(K)})$ %
is \emph{semantically} aligned with the observation. We employ ancestral sampling on our $K$-level hierarchical posterior, from $h(K)$ to $h(1)$, to generate samples $z_{h(1)}$ as in Fig. \ref{fig:vipaint-description}. 
This step gradually adds diverse image details. VIPaint then refines $z_{h(1)}$ using the prior denoising model at every step $t < h(1)$. Similar to DPS \citep{chung2023diffusion} and the Langevin dynamics~\citep{smld} underlying unconditional DM samplers, we update the samples using the gradient of the likelihood $\log p_\theta(y\mid z_t), t<h(1)$, approximated as above. This ensures fine-grained details of our final inpaintings are consistent and realistic.  By using a variational posterior to explicitly capture diffusion processes at moderate and high noise levels, and only relying on local gradient-based updates for low noise levels, our VIPaint method substantially improves on direct Langevin samplers like DPS.

%% file: sections/Experiments_Results.tex
\begin{figure*}[t] 
    \small 
  \centering
\includegraphics[width=\linewidth]{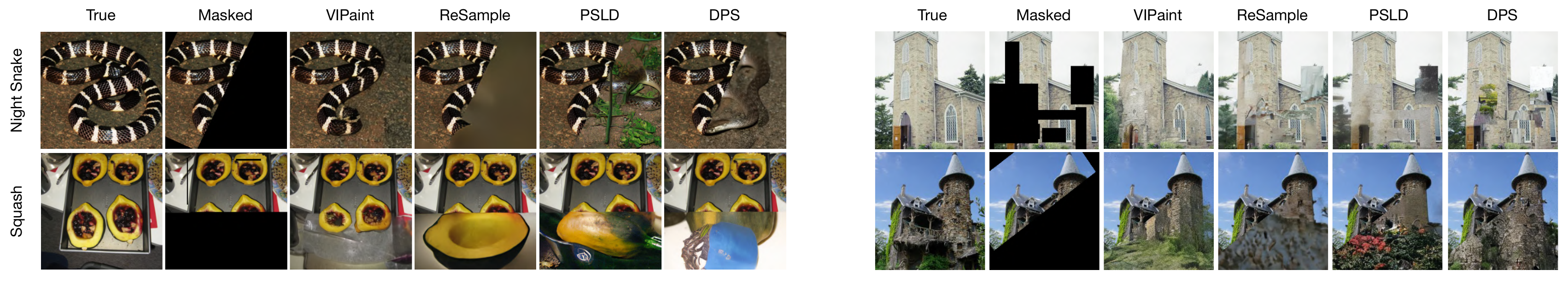}
\vspace*{-8pt}
  \caption{Inpainting with large (random or rotated window) masks using LDMs for Imagenet256 (left) and LSUN churches (right).
  DPS, PSLD, and ReSample produce blurry inpaintings. %
  Despite being conditioned on class labels,  baseline methods’ inpaintings for ImageNet are inconsistent with the observed image. In contrast, VIPaint captures global semantics, producing highly realistic inpaintings. See Appendix for more examples.
  } 
\label{fig:ldm}
\end{figure*}

\begin{table}[t]
\caption{Quantitative ImageNet64 Inpainting Results.}
\vspace*{-8pt}
\small
\centering
\resizebox{\linewidth}{!}{%
\begin{tabular}{@{}lllllllll@{}}
\toprule
           &  \multicolumn{3}{l}{Rotated Window}            & & \multicolumn{3}{l}{Random Mask}                \\ \cmidrule{2-4} \cmidrule{6-8} 
Method     & PSNR$\uparrow$          & SSIM$\uparrow$          & LPIPS$\downarrow$         & & PSNR$\uparrow$           & SSIM$\uparrow$          & LPIPS$\downarrow$         \\ \midrule
VIPaint-2  & \underline{9.24}     & \textbf{0.56}       & \textbf{0.30}         & & \textbf{13.33} & \textbf{0.62}      & \textbf{0.23} \\
CoPaint-TT & 8.51                 & 0.51                & \underline{0.32}      & & \underline{12.51}          & 0.58               & \underline{0.25}          \\
CoPaint    & 8.47                 & 0.50                & 0.35                  & & 12.12          & 0.56               & 0.28          \\
RePaint    & 8.82                 & \textbf{0.56}       & \underline{0.32}      & & 12.05          & \underline{0.59}   & 0.26          \\
DPS        & 8.15                 & \underline{0.53}    & \underline{0.32}      & & 11.45          & 0.56               & 0.29          \\
Blended    & 7.68                 & 0.52                & 0.34                  & & 11.47          & 0.57               & 0.28          \\
RedDiff    & 8.56                 & 0.45                & 0.46                  & & 11.89          & 0.51               & 0.41          \\
RedDiff-V  & \textbf{9.27}        & \underline{0.53}    & 0.41                  & & 8.35           & 0.16               & 0.67          \\ \bottomrule
\end{tabular}%
}
\caption*{
Using the pixel-based EDM prior for all methods, PSNR, SSIM, and LPIPS are averaged over 1000 inpaintings. VIPaint shows the best performance (\textbf{bold}), and the second best is \underline{underlined}.}
\vspace*{-6pt}
\label{tab:performance_metrics_pixel}
\end{table}

We first experiment with three natural image datasets that have been widely used to evaluate image restoration and inpainting with DMs:  LSUN-Church \citep{yu15lsun}, ImageNet64, and ImageNet256 \citep{deng2009imagenet}. For ImageNet64, we use the pre-trained class-conditioned pixel-space \emph{EDM} diffusion model \citep{edm}. For LSUN-Churches256 and ImageNet256, we use the pre-trained LDMs from \cite{Rombach_2022_CVPR}, as is standard in prior work. We sample (\emph{not} cherry-pick) 100 or 1000 test images for each dataset, and match our experimental settings and preprocessing to previous work (DPS, \cite{chung2023diffusion}).
We consider three masking patterns: 1) as in the easier experimental setup of REDdiff~\citep{mardani2024variational}, 1000 images with a small mask (under 30\% of image) adapted from Palette \citep{saharia2022palette};  2) 100 images using a random mask \citep{zhao2021comodgan_comodgan} covering 40-80\% of image; 3) 100 images using a randomly rotated mask covering about 50\% of image. Given the larger uncertainty for patterns 2-3, we sample and evaluate 10 reconstructions per test image (1000 reconstructions total).  We emphasize that experimental setups that mask large portions of each image provide a far more challenging inpainting benchmark.  (Note that several weak-performing baselines only evaluated on the easy, small mask scenario.)

We also demonstrate that VIPaint is applicable to the latest, foundational text-to-image diffusion models via further experiments with \emph{Stable Diffusion} (SD) 3.5 \citep{esser2024scaling}. To focus quantitative comparisons on inference performance, while avoiding potential biases from the choice of prompt, our quantitative experiments use 100 images generated from the SD 3.5 model and provide the same prompt for inpainting (see examples in Appendix). 
Some recent methods~\citep{Spagnoletti_2025_ICCV,kim2025regularization} refine the textual prompt as part of the inpainting process; incorporating prompt refinement with VIPaint is a promising direction for future research.
For qualitative comparison, we also include real images with prompts from the MSCOCO 2014 validation set \citep{lin2014microsoft}. All SD experiments use $1024 \times 1024$ images.

\begin{table}[t]
\caption{Quantitative SD3.5 Inpainting Results.}
\vspace*{-8pt}
\centering
\scriptsize
\begin{tabular}{lccc}
\toprule
Method     & PSNR$\uparrow$ & SSIM$\uparrow$ & LPIPS$\downarrow$ \\
\midrule
VIPaint-2    & \textbf{26.69} & \textbf{0.946} & \textbf{0.050} \\
DPS        & 25.22        & 0.943        & 0.061         \\
ControlNet & 22.89        & 0.703        & 0.137         \\
SD-Inpaint & 23.50        & 0.692        & 0.127        \\
\bottomrule
\end{tabular}
\caption*{Results are averaged across 10 sample completions for 100 synthetic images generated using SD3.5.}
\label{tab:performance_metrics_custom}
\end{table}

We use the notation VIPaint-$K$ to denote the number of keypoints $K$ in the hierarchical VIPaint posterior. We found empirically that discretizations and hyperparameters of VIPaint translate well between models using the same noise schedule, as demonstrated by the LSUN and ImageNet-256 latent diffusion models.

\newcommand{\STAB}[1]{\begin{tabular}{@{}c@{}}#1\end{tabular}}
\begin{table*}[t]
\caption{Quantitative $256 \times 256$ Latent DM Inpainting Results.}
\vspace{-10pt}
\centering
\scriptsize
\resizebox{\linewidth}{!}{%
\begin{tabular}{@{}llllllllllllllll@{}}
\toprule
    & \textbf{} & \multicolumn{4}{l}{Small Mask}                           &     & \multicolumn{4}{l}{Rotated Window}                       &     & \multicolumn{4}{l}{Random Mask}                               \\ \cmidrule{3-6} \cmidrule{8-11} \cmidrule{13-16}
    & Method    & PSNR$\uparrow$          & SSIM$\uparrow$          & LPIPS$\downarrow$          & KID*$\downarrow$    &     & PSNR$\uparrow$           & SSIM$\uparrow$           & LPIPS$\downarrow$         & KID*$\downarrow$    &     & PSNR$\uparrow$            & SSIM$\uparrow$           & LPIPS$\downarrow$          & KID*$\downarrow$         \\ \midrule
    \multirow{5}{*}{\STAB{\rotatebox[origin=c]{90}{Imagenet}}}
    & VIPaint-2 & 13.51          & 0.47          & 0.30          & \textbf{2.80} & & \textbf{9.43} & 0.44 & \textbf{0.39} & \textbf{5.40} & & \textbf{10.04} & \textbf{0.53} & \textbf{0.39} & \textbf{6.10} \\
    & MGPS      & \textbf{20.84} & \textbf{0.86} & \textbf{0.16} & 3.30          & & 7.89          & \textbf{0.53}          & 0.40          & 16.20         & & 9.26           & \textbf{0.53}          & \textbf{0.39}          & 25.60         \\
    & ReSample  & 15.36          & 0.58          & 0.37          & 11.00         & & 7.58          & 0.40          & 0.51          & 26.50         & & 9.57           & 0.42          & 0.48          & 28.70         \\
    & PSLD      & 14.72          & 0.52          & 0.41          & 11.00         & & 7.51          & 0.33          & 0.54          & 18.30         & & 9.50           & 0.34          & 0.52          & 16.80         \\
    & DPS       & 14.62          & 0.51          & 0.42          & 12.00         & & 7.35          & 0.32          & 0.50          & 15.90         & & 9.25           & 0.33          & 0.49          & 20.50         \\ \midrule
    \multirow{5}{*}{\STAB{\rotatebox[origin=c]{90}{LSUN  \hspace{1mm}}}}
    & VIPaint-2 & 16.18          & 0.61          & 0.20          & \textbf{0.30} & & \textbf{8.39} & 0.33          & 0.45 & \textbf{7.40} & & 9.58           & 0.34          & 0.44          & 6.60          \\
    & MGPS      & \textbf{19.02} & \textbf{0.84} & \textbf{0.16} & 4.30          & & 7.40          & 0.33          & \textbf{0.40}          & 28.60         & & \textbf{9.88}  & \textbf{0.56} & \textbf{0.36} & 24.40         \\
    & ReSample  & 17.21          & 0.64          & 0.38          & 10.90         & & 8.04          & \textbf{0.43} & 0.54          & 30.90         & & 8.95           & 0.41          & 0.56          & \textbf{6.00} \\
    & PSLD      & 13.63          & 0.43          & 0.53          & 2.40          & & 7.00          & 0.34          & 0.58          & 8.30          & & 8.33           & 0.32          & 0.58          & 7.20          \\
    & DPS       & 13.17          & 0.46          & 0.56          & 48.00           & & 7.59          & 0.32          & 0.61          & 9.80          & & 8.81           & 0.31          & 0.61          & 7.40          \\ \bottomrule
\end{tabular}%
}
\caption*{For LDMs of the ImageNet256 and LSUN-Churches256 datasets, the PSNR, SSIM, LPIPS, and KID metrics are the mean score across $1000$ inpaintings. For compactness, we report KID* = KID $\times10^3$.}
\label{tab:performance_metrics_ldm}
\end{table*}

\textbf{Baselines.} We compare VIPaint with several methods designed for pixel-based DMs: \emph{i)} blending methods, \emph{Blended} \citep{song2021scorebased} and \emph{RePaint} \citep{Lugmayr2022RePaintIU}; \emph{ii)} sampling methods, \emph{DPS} \citep{chung2023diffusion} and \emph{CoPaint} \citep{pmlr-v202-zhang23q}; \emph{iii)} the \emph{REDdiff} \citep{mardani2024variational} variational approximation. Although not exhaustive, these methods exemplify recent developments in image inpainting via DMs. For LDMs we compare VIPaint with \emph{DPS}, \emph{PSLD} \citep{rout2023solving}, \emph{MGPS} \citep{moufadvariational}, and \emph{ReSample} \citep{resample}.  %
We omit other recent methods, such as DAPS \citep{DAPS} and LATINO~\citep{Spagnoletti_2025_ICCV}, which reported reduced performance for inpainting compared to previous work. %
We report the Peak Signal-To-Noise Ratio (PSNR), Structural Similarity (SSIM,~\citet{ssim}), Kernel Inception Distance (KID,~\citet{bińkowski2018demystifying}), %
and Learned Perceptual Image Patch Similarity (LPIPS,~\citet{zhang2018perceptual}) metrics. %
We show examples of LDM inpaintings in Fig.~\ref{fig:ldm}, Fig.~\ref{fig:stable_diffusion}, and the Appendix.  %

Related work using Stable Diffusion (SD) for inverse problem inference \citep{DAPS, rout2023solving} has been limited to earlier versions (SD v1.5 and SD v2). We thus compare our SD results to our own re-implementation of DPS, and two methods that fine-tune (with substantial training cost) SD specifically for inpainting: ControlNet \citep{zhang2023adding} and SD-Inpainting \citep{esser2024scaling}.

\vspace*{-8pt}
\subsection{Image Inpainting Results}
\vspace*{-5pt}
\textbf{VIPaint enforces consistency with large masks.} The results in Tables \ref{tab:performance_metrics_pixel} and \ref{tab:performance_metrics_ldm} show that prior methods perform well for small masks, while for large masks we see a clear improvement with VIPaint. For pixel-based DMs, both RedDiff and DPS perform poorly. RePaint, CoPaint, and CoPaint-TT show relative improvements, but do not match VIPaint across any dataset or masking pattern. Notably CoPaint-TT integrates the ``time travel'' heuristic proposed by RePaint \citep{Lugmayr2022RePaintIU} with CoPaint, requiring more time than both CoPaint and VIPaint-2, but nevertheless underperforming VIPaint-2. We show imputations for multiple test examples in Fig.~\ref{fig:ldm}, %
and see that VIPaint consistently produces plausible inpaintings, while other methods fail to meaningfully inpaint large masks.  Note that previous work \citep{resample} has discussed the poor performance of PSLD. %

\begin{figure}[t] %
    \centering
    \small 
\includegraphics[width=.99\linewidth]{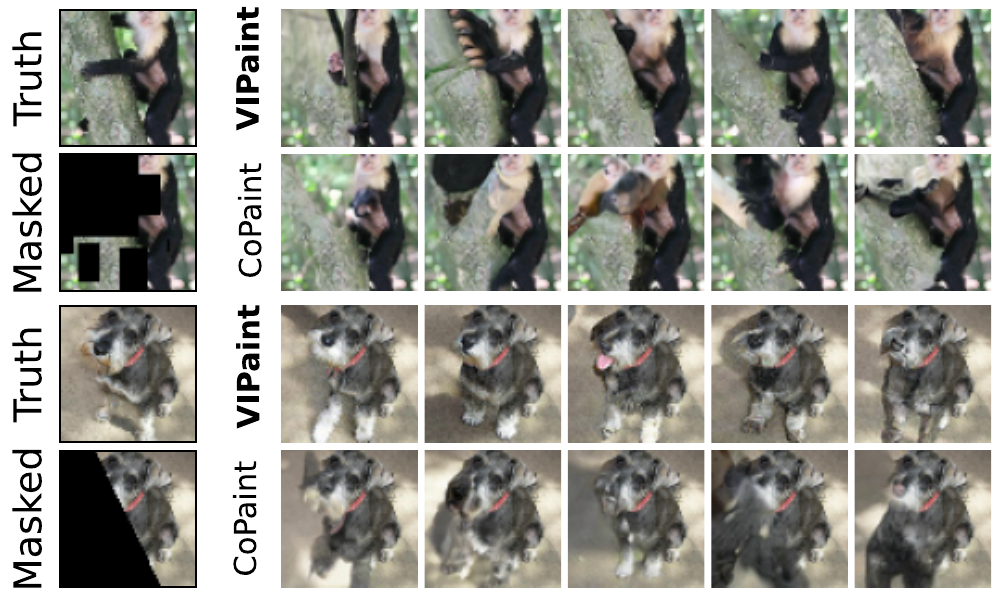}
\vspace*{-15pt}
    \caption{Sample completions comparing VIPaint with the best performing baseline, CoPaint, for two test images. We show 
    5 inpainted samples for each method. 
    VIPaint yields coherent samples while capturing uncertainty in the missing pixels in images. In contrast, CoPaint has high variance in the quality of results. 
    }
\label{fig:imagenet64-variations}
\end{figure}

\begin{figure} %
    \centering
    \small 
\includegraphics[width=\linewidth]{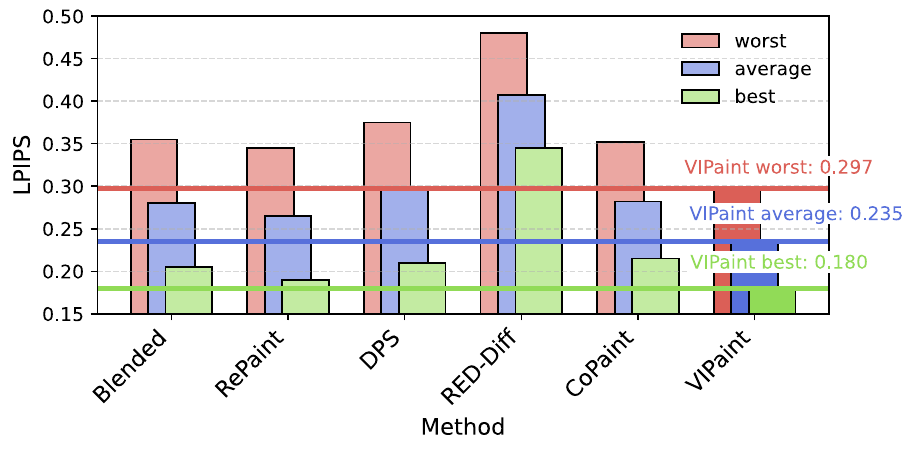}
\vspace*{-24pt}
    \caption{We compute summary statistics (minimum, mean, maximum) of the LPIPS score (lower is better) across \emph{100} sampled completions of each test image. We show the average value of each of these statistics across the full test set. We see that VIPaint improves on all baselines, both in terms of the average quality of results \emph{and} in the consistency in result quality.
    }
\label{fig:imagenet64-variations-bar}
\end{figure}

\vspace*{-2pt}
\textbf{VIPaint yields multiple plausible imputations for large masks.} We compare VIPaint with the best performing baseline, CoPaint, across multiple sample inpaintings in Fig.~\ref{fig:imagenet64-variations}. 
We observe that VIPaint produces multiple visually-plausible imputations while never violating consistency with the observed pixels. These diverse inpaintings also have better quantitative performance than baselines, see Fig.~\ref{fig:imagenet64-variations-bar}.  We show diverse imputations using different class conditioning using VIPaint in the Appendix. 

\vspace*{-1pt}
\textbf{VIPaint is effective for text-conditioned foundation models.} Results using Stable Diffusion (SD) show that our approach remains robust and effective even when integrated with state-of-the-art diffusion architectures. VIPaint consistently outperforms DPS on text-conditioned inpainting, producing images that are not only globally coherent but that also preserve fine-grained details. Moreover, VIPaint scales effectively to high-resolution $1024 \times 1024$ images. In some cases, the results exceed specialized inpainting models such as SD-Inpainting~\citep{Rombach_2022_CVPR} and ControlNet~\citep{zhang2023adding}, which were trained specifically for the inpainting task at greater computational cost. Some results can be found in Fig.~\ref{fig:stable_diffusion}, and a more comprehensive comparison is in the Appendix.

\textbf{Computational Efficiency.} Table~\ref{tab:runtime} reports the time taken by various methods to produce 10 inpaintings for one test image.  REDdiff is fast but inconsistent, and unsuitable for LDMs. Sampling methods are slower and still produce inconsistent results. VIPaint-$2$ is faster than sampling-based methods for all DMs, \emph{and} achieves better results (see Table~\ref{tab:performance_metrics_pixel}). %

\textbf{Limitations.}
VIPaint inherits biases, both good and bad, from whatever DM it is applied to.  Like other algorithms for inference with DMs, VIPaint's variational optimization has computational overhead compared to conditional models specialized to inpainting, but this is balanced by its ``zero shot'' capacity to adapt to new distortion models without expensive retraining.

\vspace*{-5pt}
\subsection{Hyperparameters}
\vspace*{-3pt}
\label{sec:hyperparameters}

\begin{figure*}[]
    \small 
    \centering 
\includegraphics[width=.8\linewidth]{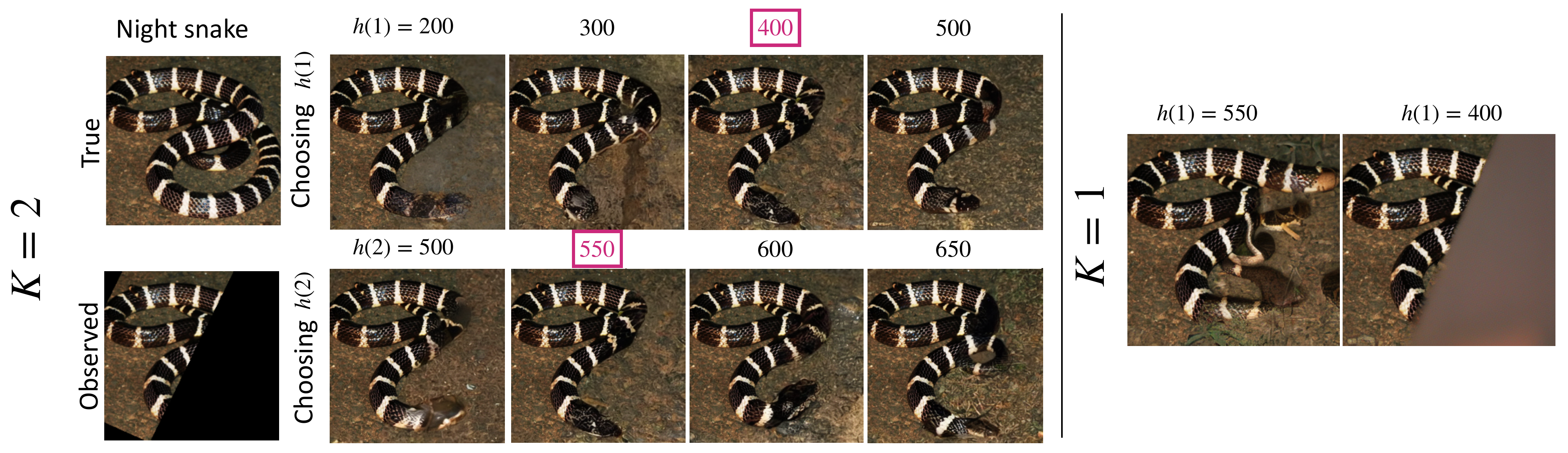}
\vspace*{-1pt}
    \caption{Sensitivity to keypoint selection for the Imagenet256 LDM. \emph{Left:} We fix either $h(1)=400$ or $h(K)=550$ and vary the other endpoint around the chosen value to demonstrate robustness. \emph{Right:} A non-hierarchical variant of VIPaint with $K=1$, for either $h(1)=400$ or $h(1)=550$, is inferior to VIPaint-2.}
    \label{fig:bottom}
\end{figure*}

\paragraph{Keypoints ($K$) in VIPaint's posterior.}
Fig.~\ref{fig:bottom} demonstrates the critical importance of using a hierarchical posterior. Removing either keypoint from the VIPaint-2 posterior results in substantially degraded results. Note that this K=1 model is similar to REDdiff, but with a variational distribution defined at $t>0$. %
Conversely, the hierarchical VIPaint-2 effectively captures both global and local details.

\begin{table}[t]
\caption{Runtime Comparison For Inference Methods.}
\vspace*{-6pt}
\scriptsize
\centering
\begin{tabular*}{\linewidth}{@{\extracolsep{\fill}} lc@{\hskip 1pt}c@{\hskip 1pt}c@{\hskip 5pt}c@{\hskip 1pt}c@{\hskip 5pt}c@{\hskip 1pt}c@{\hskip 5pt}c@{\hskip 1pt}c@{\hskip 5pt}c@{\hskip 1pt}c@{\hskip 5pt}c@{\hskip 1pt}c@{\hskip 5pt}c@{\hskip 1pt}c@{\hskip 5pt}c@{\hskip 1pt}c@{\hskip 5pt}c}
    \toprule
    Dataset & \textbf{Blended} & \textbf{DPS} & \textbf{VIPaint} & \textbf{\emph{Sample}} \\
    \midrule
   ImageNet64 &  (1.13, 1000) & (2.55, 1000) &  (1.5, 150) & (1.8, 700)  \\
    \midrule
    ImageNet256 &  (4, 1000) & (10, 500) & (2, 150)   & (8, 400) &  \\ 
    LSUN &  (1.3, 1000) & (5.1, 500)  & (2.1, 150)  &(4.3, 400)  
    \\ \bottomrule
\end{tabular*}
\caption*{The (\textit{time in minutes}, \textit{neural function evaluations}) are reported for EDM (top) and LDM (bottom) priors. 
For VIPaint, optimization (``VIPaint") and sampling are separated, since optimized posterior can be reused. 
REDdiff matches Blended, while RePaint (2.8 mins) and CoPaint (2.6 mins) are slightly slower than DPS.}
\label{tab:runtime}
\end{table}

We find that increasing the number of keypoints in the hierarchy typically improves results for more challenging settings, such as large-mask inpainting on Imagenet256. In this case, VIPaint-4 $(K=4)$ improves average LPIPS from 0.392 to 0.358 for rotated window masks, and from 0.409 to 0.373 for random masks, relative to VIPaint-2. %
Increasing $K$ has quickly diminishing returns as each update becomes more expensive and optimization becomes more complex, requiring more iterations to converge. For $K>2$, we find that up-weighting the KL-divergence terms of the loss, similar to the $\beta$-VAE \citep{higgins2017betavae}, can help speed and stabilize convergence while encouraging solutions with higher variance. In general, we recommend $K=2$ for most settings, as this $2$-level posterior can be optimized in as low as $50$ iterations and tackles the difficult problem of large mask inpainting well.

\vspace*{-8pt}
\paragraph{Choosing $h(1), h(K)$.} We chose the endpoints of VIPaint's posterior based on qualitative analysis on a few validation images, and fixed these values for all experiments (except where noted). We found that spreading keypoints across signal-to-noise ratios ($\alpha_t^2/\sigma_t^2) \in [0.2, 0.5]$ led to good results across models and datasets, concentrating posterior inference on the noise levels which are most crucial to perceptual image quality. 
The Appendix has further details, and Fig.~\ref{fig:bottom} illustrates robustness to this hyperparameter.

\subsection{Deblurring and Superresolution Results}

\begin{figure}
    \small 
    \centering 
\includegraphics[width=\linewidth]{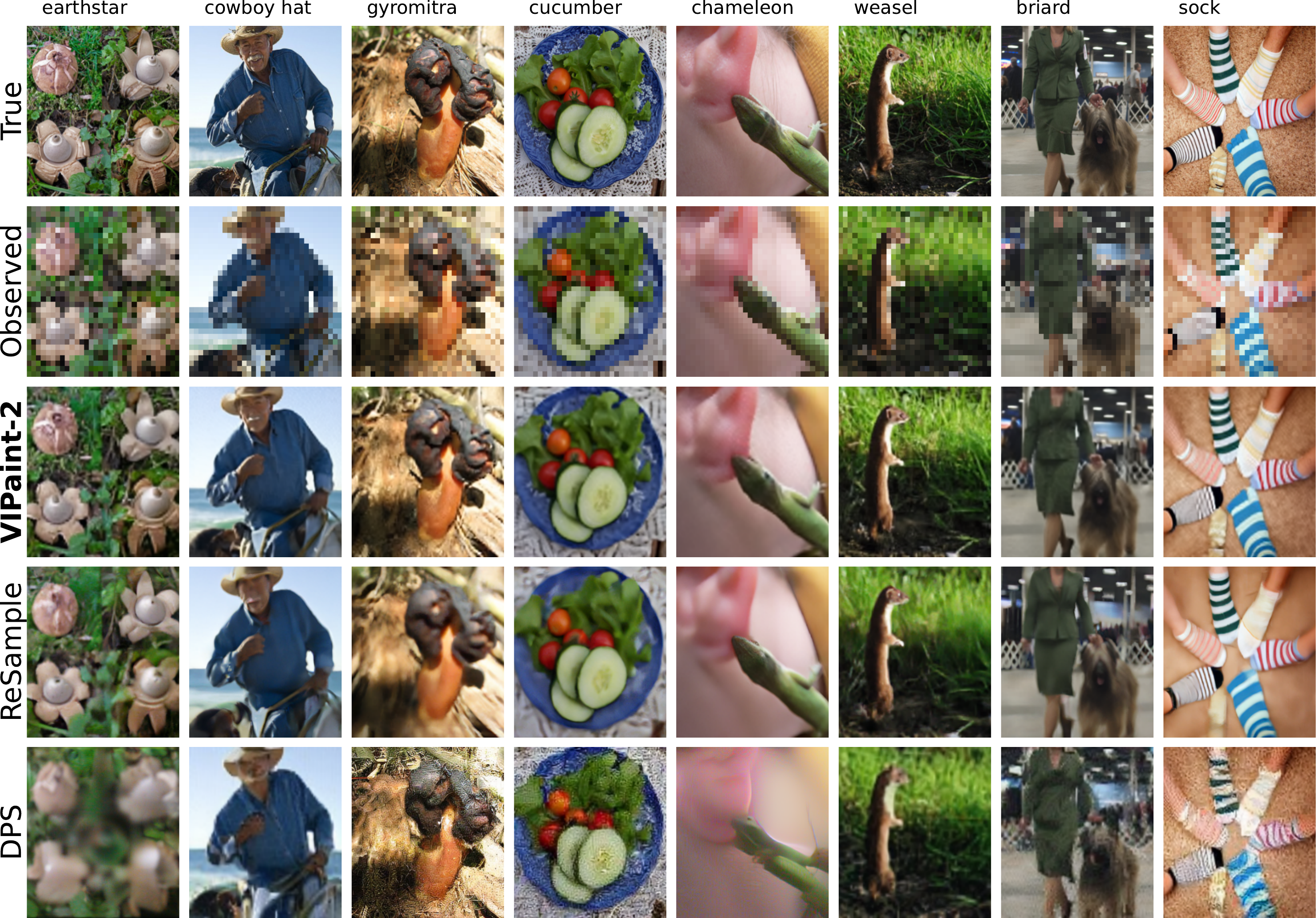}
\vspace*{-15pt}
    \caption{Qualitative examples of superresolution reconstruction with Imagenet256 LDMs. 
We see that DPS produces extremely blurry images, an artifact that ReSample only partially improves. In contrast, VIPaint-2 leads to samples closer to the true image and produces \emph{very} realistic images.}
    \label{fig:super-resolution2}
\end{figure}

While VIPaint is primarily motivated by the challenge of inpainting large regions with diffusion models, it is a general-purpose inference algorithm applicable to other inverse problems. %
We consider two linear inverse problems that are widely used as benchmarks, Gaussian deblurring and superresolution. For Gaussian deblurring, we use a kernel with size 61 × 61 with standard deviation 3.0. For superresolution, we use bicubic downsampling, and a similar experimental setup as \citet{chung2023diffusion}. 

\begin{figure}
    \small 
    \centering 
\includegraphics[width=\linewidth]{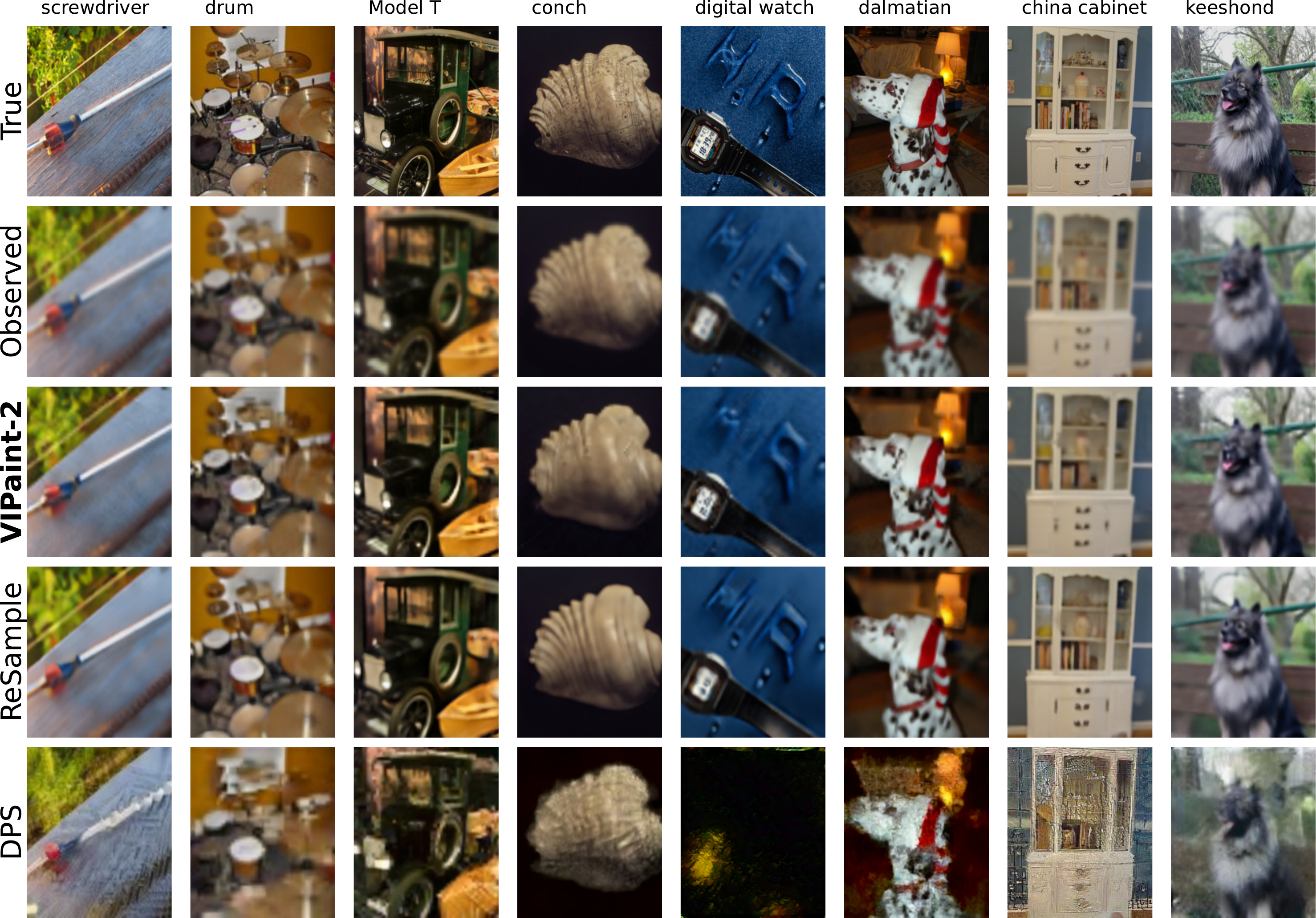}
\vspace*{-15pt}
    \caption{Qualitative examples of removing Gaussian blur with Imagenet256 LDMs.  VIPaint again produces sharper results with fewer artifacts than baselines methods.}
    \label{fig:gaussian-deblurring2}
\end{figure}

We compare the performance of VIPaint with ReSample, PSLD, and DPS for ImageNet256 LDM prior.  For the pixel-based ImageNet64 EDM model, we compare the VIPaint Gaussian Deblurring performance to DPS. We report the PSNR and  LPIPS metrics in Table \ref{tab:performance_metrics_LIP_}, and give examples in Fig.~\ref{fig:super-resolution2},~\ref{fig:gaussian-deblurring2} and the Appendix. 
We see that for complex image datasets like ImageNet, VIPaint shows strong advantages over DPS and PSLD, and performs similarly to ReSample.

\begin{table}[t]
\caption{Superresolution and Deblurring Results.}
\label{tab:performance_metrics_LIP_}
\vspace*{-6pt}
\small
    \centering
     \resizebox{\linewidth}{!}{%
    \begin{tabular}{lc@{\hskip 1pt}c@{\hskip 5pt}c@{\hskip 1pt}c@{\hskip 6pt}c@{\hskip 1pt}c@{\hskip 5pt}c@{\hskip 1pt}c@{\hskip 5pt}c@{\hskip 1pt}c@{\hskip 5pt}ccccc}
    \toprule
    & \multicolumn{4}{c}{ImageNet256} & \multicolumn{2}{c}{ImageNet64} \\ 
    Task  & \multicolumn{2}{c}{Super-resolution 4x} & \multicolumn{2}{c}{Gaussian Deblur} & \multicolumn{2}{c}{Gaussian Deblur} \\ 
    \midrule
     Metric & LPIPS $\downarrow$ & PSNR $\uparrow$& LPIPS $\downarrow$ & PSNR $\uparrow$ & LPIPS $\downarrow$ & PSNR $\uparrow$ \\
    \midrule
    VIPaint & \textbf{0.37} &  \textbf{18.90}  & \underline{0.45}  & \underline{17.91} & \textbf{0.31} &  \textbf{13.60}  \\
    ReSample & \underline{0.40} & \underline{18.41} & \textbf{0.44} & \textbf{18.03} & -- & --  \\
PSLD & 0.67 & 7.77 & 0.58 & 0.02 & -- & --\\
DPS  &  0.58 & 12.99  & 0.60 & 12.61 & \underline{0.32}  &  \underline{13.43}  \\
\bottomrule
    \end{tabular}
    }
    \caption*{Quantitative results (LPIPS, PSNR) for solving linear inverse problems on ImageNet256 using LDM priors, and ImageNet64 using EDM priors. Best results are in bold, and second best results are underlined.} 
\end{table}

%% file: sections/Conclusion.tex
VIPaint is a principled and general approach to adapt pretrained DMs for image inpainting and other inverse problems. We take widely used (latent) diffusion generative models, allocate variational parameters for the latent codes of each partial observation, and fit the parameters stochastically to optimize the induced variational bound. The simple but flexible structure of our bounds allows efficient VIPaint optimization to outperform previous sampling and variational methods, even for high-resolution text-conditioned LDMs.

%% file: sections/appendix.tex
\section{Additional Results}
\label{app:addn_results}

\begin{figure}[p] 
    \small 
  \centering
\includegraphics[width=\linewidth]{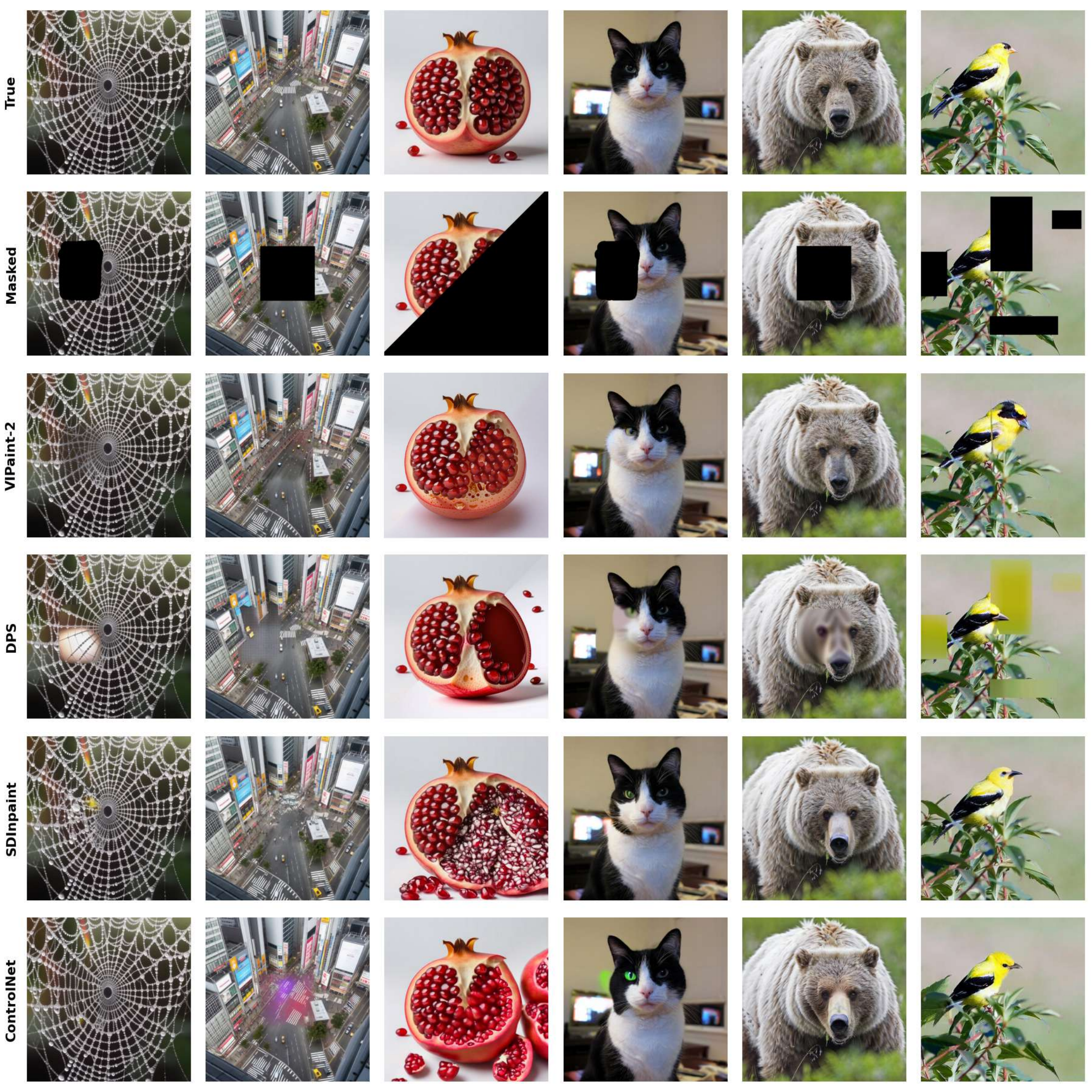}
\vspace*{-10pt}
  \caption{This figure shows image completion results using the LDM prior with the pretrained Stable Diffusion 3.5 turbo model, conditioned on image prompts. Results are reported for synthetic data (left three) and MSCOCO Validation 2014 (right three) under different inpainting masks (Drawn Mask, Center Mask, and Random Masking). While DPS produces blurry reconstructions and the baseline method often fails to remain consistent with the observed image, VIPaint 2.0 successfully captures global semantics and produces realistic inpaintings. Example prompts include: “Close-up of dew-covered spiderweb, rainbow refractions in droplets,” “Night cityscape of Tokyo’s Shibuya Crossing from rooftop, neon lights on wet pavement,” “Close-up of freshly cut pomegranate with glistening seeds,” “A black-and-white tuxedo cat with bright green eyes sitting upright indoors, looking directly at the camera in front of a blurred background,” “A large brown bear with thick fur walking forward through tall green grass, staring straight ahead,” and “A small bright yellow bird with black wings perched on leafy green branches in sunlight, with a soft blurred background.” 
  } 
\label{fig:sd35}
\end{figure} 

\begin{figure}[p] 
    \small 
  \centering
\includegraphics[width=\linewidth]{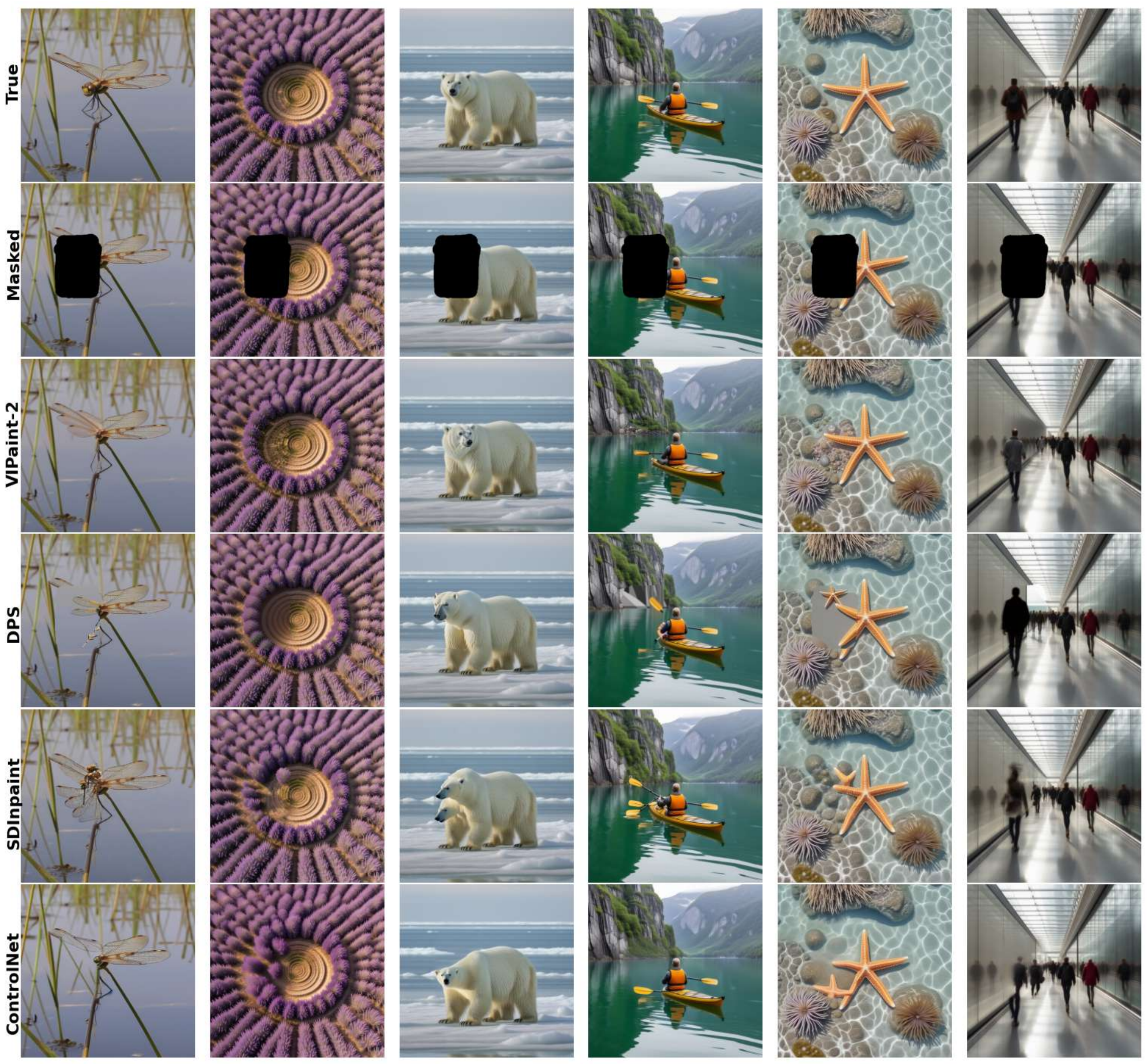}
\vspace*{-10pt}
  \caption{This figure shows image completion results using the LDM prior with the pretrained Stable Diffusion 3.5 turbo model, conditioned on image prompts. Results are reported for synthetic data drawn masks. While DPS produces blurry reconstructions and the baseline method often fails to remain consistent with the observed image, VIPaint 2.0 successfully captures global semantics and produces realistic inpaintings. Prompts used include: “Dragonfly perched on reed above still pond, wings catching light,” “Drone shot of a perfectly circular lavender field under the midday sun,” “Polar bear standing on sea ice, vast Arctic horizon behind,” “Kayaker paddling through a still emerald fjord in Norway, cliffs reflected,” “Shallow tide pool with starfish and anemones, clear water surface,” and “Long exposure of commuters moving through a glass atrium, ghostly trails.”
  } 
\label{fig:sd35_2}
\end{figure} 

\subsection{Conditional Inpainting: Text Conditioned}
Using Stable Diffusion 3.5~\citep{esser2024scaling}, our method remains effective when integrated with state-of-the-art diffusion architectures and scales to high-resolution 1024×1024 images. For text-conditioned inpainting, VIPaint consistently outperforms DPS, yielding images that maintain global coherence and preserve fine details. In several cases, VIPaint also surpasses task-specific inpainting models, including SD-Inpainting \citep{esser2024scaling} and ControlNet \citep{zhang2023adding}. Minor artifacts remain in baseline methods, such as boundary blending inconsistencies and repetitive filling of masked regions from the prompt without considering the surrounding context, leading to globally inconsistent results. See Figs.~\ref{fig:sd35},~\ref{fig:sd35_2} for qualitative examples.

Synthetic images are generated using Stable Diffusion~3.5 at a resolution of $1024\times1024$. 
The text prompts used for image generation were produced by ChatGPT using the instruction: 
``Generate 100 realistic image-generation prompts. Each prompt should describe a single, coherent scene in natural language with photographic details (camera type, lighting, time of day, composition, mood, and subject). Focus on professional photography terms (e.g., long exposure, shallow depth of field, aerial view, macro, golden hour).''

\subsection{Linear Inverse Problems}
\label{app:linear-inverse-problems}

\begin{figure} \centerline{\includegraphics[width=0.8\linewidth]{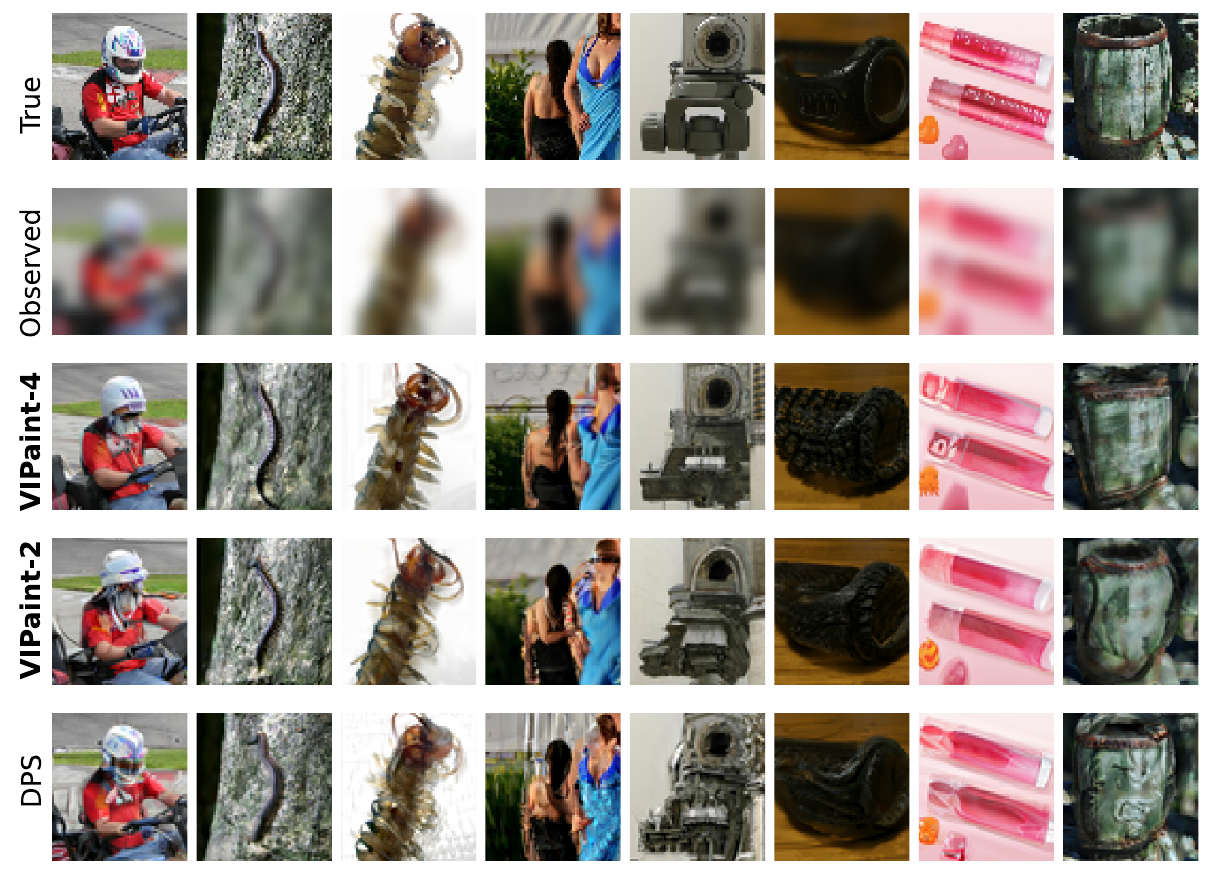}}
  \caption[]
{\small Qualitative results for Gaussian DeBlurring using EDM prior for ImageNet64. We see  VIPaint leads to samples closer to the true image and produces \emph{more} realistic images.
} 
\label{fig:imagenet64_gd}
\end{figure}

To evaluate VIPaint performance at linear inverse problems other than inpainting, we also consider Gaussian deblurring and superresolution. 
We compare the performance of VIPaint with ReSample, PSLD \& DPS for ImageNet256 dataset using the LDM prior. For the pixel-based model, we include results for Gaussian Deblurring comparing VIPaint with DPS. 
Some qualitative plots are in Fig.~\ref{fig:super-resolution2},~\ref{fig:gaussian-deblurring2} and \ref{fig:imagenet64_gd}.

\subsection{Conditional Inpainting}

\begin{figure} 
 \centering
\centerline{\includegraphics[width=\linewidth]{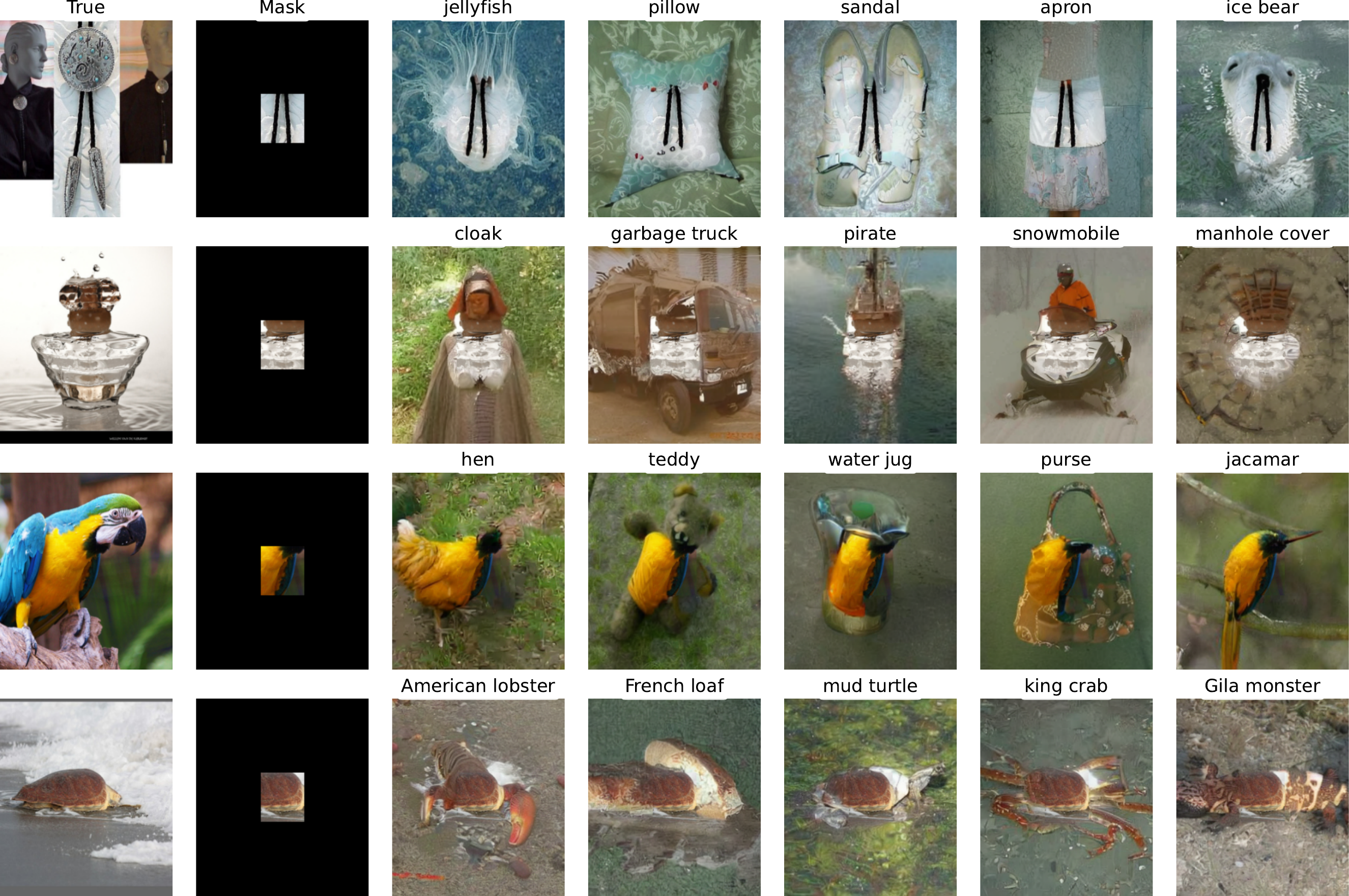}}
  \caption[]
{\small Qualitative results for VIPaint diversity for ImageNet256 with LDM prior using different class conditioning. We see that VIPaint follows the input label and ensures consistency with the observed set of pixels.
} 
\label{fig:variety-imagenet256}
\end{figure}

For the case of large-mask image inpainting, we perform some qualitative experiments where we change the input class condition for a given masked observation into the diffusion generative model as shown in Fig \ref{fig:variety-imagenet256}. We see that VIPaint generates images consistent with the different input class condition \emph{while} also enforcing consistency with the observed set of pixels.  

\newpage
\section{Diffusion Models: Definition \& Training Procedure Recap }
\label{app:diff_recap}
\subsection{Forward time Diffusion Process} 
\label{app:forward-time}
The background and expressions on forward diffusion process is taken from \cite{vdms} and included here for completeness. Re-iterating $q(z_t \mid \bar{x}=\textbf{enc}(x)) = \mathcal{N}(z_t \mid \alpha_t \bar x, \sigma_t^2 I), $ we have the forward diffusion as: 
\begin{equation}
    q(z_t \mid x) = \mathcal{N}(\alpha_t x , \sigma_t^2 I).
\end{equation}
\paragraph{Forward Conditional $q(z_t | z_s)$: }
The distribution $q(z_t | z_s)$ for any $t>s$ are also Gaussian, and from \cite{vdms}, we can re-write as 
\begin{align}
    & \mathcal{N}(\alpha_{t|s}z_s, \sigma^2_{t|s} I) \\
    \text{where,} \quad  & \alpha_{t|s} = \alpha_t/\alpha_s, \\
     \text{and,} \quad  & \sigma_{t|s}^2 = \sigma_t^2 - \alpha_{t|s}^2 \sigma_s^2
\end{align}
\paragraph{Reverse Conditional, $q(z_s | z_t, x)$:}
The posterior $q(z_s|z_t, x)$ from \cite{vdms} can be written as: 
\begin{align}
    q(z_s | z_t, x) = \mathcal{N}(\mu_Q(z_t, x; s,t), \sigma_Q^2(s,t) I) \\
    \text{where,} \quad \sigma_Q^2(s,t) = \sigma_{t|s}^2 \sigma_s^2 / \sigma_t^2 \\
    \text{and,} \quad \mu_Q(z_t, x; s,t) = \frac{\alpha_{t|s}\sigma_s^2}{\sigma_t^2} z_t+ \frac{\alpha_s\sigma_{t|s}^2}{\sigma_t^2}x
\end{align}
\subsection{Reverse Diffusion: Defining \texorpdfstring{$p_\theta(z_s | z_t)$}{}}
\label{app:reverse-diffusion}
Here, we describe in detail the conditional reverse model distributions $p_\theta(z_s | z_t)$ for the two cases of variance-exploding and variance preserving diffusion process. Given these formulations, it is straightforward to compute the KL distance between our posterior $q_{\lambda}(z_s | z_t, y)$ and the prior $p_\theta(z_s | z_t)$ in our loss objective (Eq. \ref{eq:VIPaintobj}) since both are conditionally Gaussian distributions and computing the KL between two Gaussians can be done in closed form.

\paragraph{Variance Exploding Diffusion Process.}
In this case, $\alpha_t = 1$ and $\sigma_t$ is usually in the range $[0.002, 50]$ \cite{song2021scorebased}. We follow the ancestral sampling rule from the same work to define our prior conditional Gaussian distributions $p_\theta(z_s | z_t)$ : 
\begin{align}
    p_\theta(z_s | z_t) = \mathcal{N}(\mu_\theta( z_t; s,t), \sigma_Q^2(s,t) I) \\
    \label{eq:VE_std}
    \text{where,} \quad \sigma_Q^2(s,t) =  (\sigma_t^2 - \sigma_s^2)\frac{\sigma_s^2}{\sigma_t^2}  \\
    \text{and,} \quad \mu_\theta(z_t; s,t) =  \frac{\sigma_s^2}{\sigma_t^2}z_t + \frac{\sigma_t^2 - \sigma_s^2}{\sigma_t^2} \hat x_\theta(z_t, t) 
\end{align}
where $\hat x_\theta(z_t, t) =   z_t - \sqrt{(\sigma_t^2 - \sigma_s^2)} * \epsilon_\theta(z_t,t)$

\paragraph{Variance Preserving Diffusion Process.}
In this case, $\alpha_t = \sqrt{1 - \sigma_t^2}$ and $\sigma_t^2$ is usually in the range $[0.001, 1]$ \cite{hoddpm}. We follow the DDIM sampling rule \cite{ddim} to define our prior conditional Gaussian distributions $p_\theta(z_s | z_t)$. This sampling rule is widely used to generate unconditional samples in small number of steps, and naturally becomes a key design choice of our prior. Here, 
\begin{align}
    p_\theta(z_s | z_t) = \mathcal{N}(\mu_\theta( z_t; s,t), \sigma_Q^2(s,t) I) \\
     \label{eq:VP_std}
    \text{where,} \quad \sigma_Q^2(s,t) = \eta (\frac{1 - \alpha_{t-1}}{1 - \alpha_t}) (1 - \frac{\alpha_t}{\alpha_{t-1}}) \\
    \text{and,} \quad \mu_\theta(z_t; s,t) =  \sqrt{\alpha_{t-1}} \hat x_\theta(z_t, t) + \sqrt{1 - \alpha_t - \sigma_t^2} \epsilon_\theta(z_t, t)
\end{align}
where $\hat x_\theta(z_t, t) = \frac{z_t - \sqrt{1 - \alpha_t} \epsilon_\theta(z_t, t)}{\sqrt{\alpha_t}}$. This schedule is adopted by the Latent Diffusion models.

\paragraph{Rectified Flow.}
In this case, $\alpha_t = 1 - t$ and $\sigma_t = t$, corresponding to a linear interpolation between data and noise ~\citep{liu2022flow}. The forward process is given by $z_t = (1-t) x + t \epsilon$, where $\epsilon \sim \mathcal{N}(0, I)$. The model is trained to predict the velocity field $v_\theta(z_t, t) \approx \epsilon - x$, from which we can recover the clean data estimate as $\hat x_\theta(z_t, t) = z_t - t \cdot v_\theta(z_t, t)$. We follow the Euler sampling rule to define our prior conditional distributions $p_\theta(z_s | z_t)$:
\begin{align}
    p_\theta(z_s | z_t) &= \mathcal{N}(\mu_\theta(z_t; s, t),\; \sigma_Q^2(s,t) I) \\
    \label{eq:RF_std}
    \text{where,} \quad \sigma_Q^2(s,t) &= 0 \\
    \text{and,} \quad \mu_\theta(z_t; s,t) &= z_t + (s - t) \cdot v_\theta(z_t, t)
\end{align}
which corresponds to a deterministic ODE step.

\subsection{Derivation of Objective for training Diffusion Models: \texorpdfstring{$\mathcal{L}_{(0,T)}(z_0)$}{} }
\label{section:derivation-dm}
We begin with the standard formulation of the negative ELBO:
\begin{align*}
    -\log p_\theta(x) &\leq  \mathcal{L}(\theta; x) = \mathbb{E}_{z_{0:T}}\left[\log \frac{q(z_{0:T}\mid x)}{p(x\mid z_0)p_\theta(z_{0:T})} \right] \\
    &= \mathbb{E}_{z_{0:T}}\left[
        \sum_{t=1}^{T}\log \frac{q(z_{t-1}\mid z_{t}, x)}{p_\theta(z_{t-1}\mid z_{t})}
        + \log \frac{q(z_T \mid x)}{p(z_T)}
        -\log p(x\mid z_0)
    \right]
    \end{align*}

As $p(z_T)\approx q(x_T\mid x) \equiv \mathcal{N}(0, I)$, and $p(x\mid z_0)$ are all fixed, we may consider them a constant for the purposes of optimizing $\theta$
    
\begin{align*}
    &= \sum_{t=1}^{T} \mathbb{E}_{z_{t}, z_{t-1}}\left[
        \log \frac{q(z_{t-1}\mid z_{t}, x)}{p_\theta(z_{t-1}\mid z_{t})}
    \right] + C \\
&= T~\mathbb{E}_{t, z_t}\left[ D_{KL}\left[q(z_{t-1}\mid z_{t}, x)~||~p_\theta(z_{t-1}\mid z_{t}) \right]\right] +C
\end{align*}

Recall that
\begin{align}
    p_\theta(z_{t-1} \mid z_t) = q(z_{t-1} \mid z_t, \bar x = \hat x_\theta( z_t,t)), 
   \text{where} \quad \hat x_\theta( z_t, t) = \frac{z_t - \sigma_t \hat \epsilon_\theta(z_t,t)}{\alpha_{t}}.
\label{eq:one-step-denoising_b}
\end{align}

By definition, these two distributions have equal variance ($\sigma^2_{Q}(t-1,t)$), differing only by the means, $\mu_{Q}(t-1,t)$ and $\mu_{\theta}(t-1,t)$, as define above. Therefore we can re-write the KL-divergence as:
\begin{align*}
D_{KL} 
    &= \frac{1}{2\sigma^2_{Q}(t-1,t)}||\mu_{Q}(z_t;t-1,t) - \mu_{\theta}(t-1,t)||^2_2 \\
    &= \frac{1}{2}\frac{\sigma^2_t}{\sigma^2_{t|t-1}\sigma^2_{t-1}}||\left(\frac{\alpha_{t|t-1}\sigma_{t-1}^2}{\sigma_t^2} z_t+ \frac{\alpha_{t-1}\sigma_{t|t-1}^2}{\sigma_t^2}x\right) - \left(\frac{\alpha_{t|t-1}\sigma_{t-1}^2}{\sigma_t^2} z_t+ \frac{\alpha_{t-1}\sigma_{t|t-1}^2}{\sigma_t^2}\hat{x}_\theta(z_t, t)\right)||^2_2\\
    &= \frac{1}{2}\frac{\sigma^2_t}{\sigma^2_{t|t-1}\sigma^2_{t-1}}||\frac{\alpha_{t-1}\sigma_{t|t-1}^2}{\sigma_t^2} \left( x - \hat{x}_\theta(z_t, t)\right)||^2_2 \\
    &= \frac{1}{2}\frac{\sigma^2_{t|t-1}}{\sigma^2_t\sigma^2_{t-1}}|| x - \hat{x}_\theta(z_t, t)||^2_2 \\
    &= \frac{1}{2}\frac{\sigma^2_{t|t-1}}{\sigma^2_t\sigma^2_{t-1}}||\frac{1}{\alpha_t}(z_t - \sigma_t\epsilon)  - \frac{1}{\alpha_t}(z_t - \sigma_t\epsilon_\theta(z_t;t))||^2_2 \\
    &= \frac{1}{2}\frac{\sigma^2_{t|t-1}}{\sigma^2_t\sigma^2_{t-1}}||\frac{\sigma_t}{\alpha_t}(\epsilon -\epsilon_\theta(z_t;t))||^2_2 \\
    &= \frac{1}{2}\left( \frac{\sigma_t^2\alpha_{t-1}^2}{\alpha_{t}^2\sigma_{t-1}^2} - 1\right)||\epsilon -\epsilon_\theta(z_t;t)||^2_2 \\
\end{align*}

Substituting back into the negative ELBO formulation, we get:
\begin{align}
     -\log p_\theta(x) \leq \mathcal{L}(\theta; x) &= \frac{T}{2} \mathbb{E}_{t,\epsilon}\bigg[\left( \frac{\sigma_t^2\alpha_{t-1}^2}{\alpha_{t}^2\sigma_{t-1}^2} - 1\right)||\epsilon - \hat \epsilon_\theta(\alpha_t\bar x+\sigma_t\epsilon,t) ||_2^2\bigg] + C.
\label{eq:kl2}
\end{align}

\subsection{Existing Posterior Sampling Methods for Inverse Problems}
\label{app:sampling-methods}
\textbf{Blending Methods} methods \citep{song2022solving, wang2023zeroshot} define a procedural, heuristic approximation to the posterior and is tailored for image inpainting. They first generate unconditional samples $z_{t-1}$ from the prior using the learned noise prediction network, and then incorporate $y$ by replacing the corresponding dimensions with the observed measurements. RePaint \citep{Lugmayr2022RePaintIU}  attempts to reduce visual inconsistencies caused by blending via a resampling strategy. A “time travel” operation is introduced, where images from the current time step $z_{t-1}$ are first blended with the noisy version of the observed image $y_{t-1}$, and then used to generate images in the $(t-1) + r, (r \geq 1)$  time step by applying a one-step forward process and following the Blended denoising process. 

\vspace*{-8pt}
\paragraph{Gradient-Based Methods.} 
Motivated by the goal of addressing more general inverse problems, Diffusion Posterior Sampling (\emph{DPS}) \citep{chung2023diffusion} uses Bayes' Rule to sample from $p_\theta(z_{t-1} | z_{t}, y) \propto  p_\theta(z_{t-1} | z_{t}) p_\theta(y | z_{t-1})$. Instead of directly blending or replacing images with noisy versions of the observation, DPS uses the gradient of the likelihood $\log p_\theta(y|z_t)$ to guide the generative process at every denoising step $t$. Since computing $\nabla_{z_t} \log p(y | z_{t-1})$ is intractable due to the integral over all possible configurations of $z_{t'}$ for $t' < t-1$, DPS  approximates $p(y | z_{t-1})$ using a one-step denoised prediction $\hat x$ using Eq.~\eqref{eq:one-step-denoising_b}. The likelihood $p(y|x) = \mathcal{N}(f(x), \sigma_v^2)$ can then be evaluated using these approximate predictions. To obtain the gradient of the likelihood term, DPS require backpropagating gradients through the denoising network used to predict $\hat x$. 

Specializing to image inpainting, \emph{CoPaint} \citep{pmlr-v202-zhang23q} augments the likelihood with another regularization term to generate samples $z_{t-1}$ that prevent taking large update steps away from the previous sample $z_{t}$, in an attempt to produce more coherent images. Further, it proposes CoPaint-TT, which additionally uses the time-travel trick to reduce discontinuities in sampled images.

Originally designed for pixel-space diffusion models, it is difficult to adopt these works directly to latent diffusion models. Posterior Sampling with Latent Diffusion (\emph{PSLD}) \citep{rout2023solving} first showed that employing \emph{DPS} directly on latent space diffusion models produces blurry images. It proposes to add another ``gluing'' term to the measurement likelihood which penalizes samples $z_{t}$ that do not lie in the encoder-decoder shared embedding space. However, this may produce artifacts in the presence of measurement noise (see \cite{resample}). To address this issue, recent concurrent work on the \emph{ReSample} \citep{resample} method divides the timesteps in the latent space into 3 subspaces, and optimizes samples $z_t$ in the mid-subspace to encourage samples that are more consistent with observations. Other work \citep{yu2023freedom} highlights a 3-stage approach where data consistency can be enforced in the latter 2 stages which are closer to $t=0$.

\newpage
\section{VIPaint: VI method using Diffusion Models as priors}
\label{app:vipaint_math}
\subsection{Derivation of VIPaint's Training Objective}

As specified in the main paper, we define a variational distribution over the latent space variable $z$ as $q_\lambda(z)$ and re-use the diffusion prior to generate $x \sim p_\theta(x \mid z)$. We derive the variational objective here: 
\begin{align}
&\mathcal{L}(\lambda; y) \nonumber \\
&= \mathbb{E}_{q_\lambda(z,x)}[\log p_\theta(y, x, z) -  \log q_\lambda(z, x \mid y)] \nonumber \\
&= \mathbb{E}_{q_\lambda(z,x)}[\log p_\theta(z) + \log p_\theta(x \mid z_{h(1)}) + \log p_\theta(y \mid z_{h(1)}) -  \log q_\lambda(z) - \log q_\lambda(x \mid z_{h(1)})] \nonumber \\
&= \mathbb{E}_{q_\lambda(z)}[\log p_\theta(y \mid z_{h(1)}) + \log p_\theta(z) -  \log q_\lambda(z)] \nonumber \\
&= \mathbb{E}_{q_\lambda(z)}[\log p_\theta(y \mid z_{h(1)})] - \mathbb{E}_{q_\lambda(z)}[\log q_\lambda(z) - \log p_\theta(z) ] \nonumber \\
&= \mathbb{E}_{q_\lambda(z)}[\log p_\theta(y \mid z_{h(1)})] -  \underbrace{\mathbb{E}_{q_\lambda(z)}[\log q_\lambda(z) - \log p_\theta(z) ]}_\text{second term} 
\end{align}
The second term can be further decomposed as: 
\begin{align}
&= \mathbb{E}_{q_\lambda(z)}[\sum_{i=1}^{K-1}\log q_\lambda(z_{h(i)} \mid z_{h(i+1)}) + \log q_\lambda(z_{h(K)}) - \sum_{i=1}^{K-1}\log p_\theta(z_{h(i)} \mid z_{h(i+1)}) - \log p_\theta(z_{h(K)}) ] \nonumber \\
&= \sum_{i=1}^{K-1} \mathbb{E}_{z_{h(i+1)}} D[q_\lambda(z_{h(i)} \mid z_{h(i+1)}) || p_\theta(z_{h(i)} \mid z_{h(i+1)})] - \underbrace{D(q(z_{h(K)}) || p(z_{z_{h(K)}}))}_\text{diffusion loss}
\end{align}
Finally, $\mathcal{L}(\lambda; y) \nonumber$
\begin{align}
&= \mathbb{E}_{q_\lambda(z)}[\log p_\theta(y \mid z_{h(1)})] - \sum_{i=1}^{K-1}  \mathbb{E}_{z_{h(i+1)}}  D[q_\lambda(z_{h(i)} \mid z_{h(i+1)}) || p_\theta(z_{h(i)} \mid z_{h(i+1)})] - \underbrace{D(q(z_{h(K)}) || p(z_{z_{h(K)}}))}_\text{diffusion loss}
\end{align}

Negating the above objective, we get the loss objective as follows.
\begin{align*}
    L(\lambda) = &\underbrace{- \mathbb{E}_{q}[\log p_\theta(y\mid z_{h(1)})]}_\text{reconstruction loss} + \beta \underbrace{ \mathcal{L}_{(h(K),T)}(z_{h(K)})}_\text{diffusion loss} + \beta \underbrace{\sum_{i=1}^{K-1} \mathbb{E}_{z_{h(i+1)}}D\Big[ q_\lambda(z_{h(i)} \mid z_{h(i+1)})~||~p_\theta(z_{h(i)} \mid z_{h(i+1)} )) \Big]}_\text{hierarchical loss}.
\label{eq:VIPaintobjApp}
\end{align*}

Now, let's derive the diffusion loss term %
in the following subsection. %

\subsection{Derivation of Diffusion Loss for VIPaint} 
\label{section:derivation-diffloss}
For any $ h(K) < s< t< T$, we have : 
\begin{align}
    & \mathbb{E}_{z_{h(K)} \sim q_\lambda(z_{h(K)})} \Bigg[  \log \frac{q(z_{h(K)+1:T}|z_{h(K)})}{p_\theta(z_{h(K):T})} \Bigg ] \\
    &= \mathbb{E}_{z_{h(K)} \sim q_\lambda(z_{h(K)})} \Bigg[  - \log p(z_T) + \sum_{t \geq h(K)} \log \frac{q(z_{t}|z_{s})}{p_\theta(z_{s}|z_t)} \Bigg ]\\  
    &= \mathbb{E}_{z_{h(K)} \sim q_\lambda(z_{h(K)})} \Bigg[  - \log p(z_T) + \sum_{t > h(K)} \log \frac{q(z_{t}|z_{s})}{p_\theta(z_{s}|z_t)} + \log \frac{q(z_{h(K)+1}|z_{h(K)})}{p_\theta(z_{h(K)} | z_{h(K)+1})} \Bigg ]\\  
    &= \mathbb{E}_{z_{h(K)} \sim q_\lambda(z_{h(K)})} \Bigg[  - \log p(z_T) + \sum_{t > h(K)} \log \frac{q(z_{s}|z_{t}, z_{h(K)})}{p_\theta(z_{s}|z_t)} \cdot \frac{q(z_{t}|z_{h(K)})}{q(z_{s} |z_{h(K)})}  + \log \frac{q(z_{h(K)+1}|z_{h(K)})}{p_\theta(z_{h(K)} | z_{h(K)+1})} \Bigg ]\\  
    &= \mathbb{E}_{z_{h(K)} \sim q_\lambda(z_{h(K)})} \Bigg[  - \log \frac{p(z_T)}{q(z_T | z_{h(K)})} + \underbrace{\sum_{t > h(K)} \log \frac{q(z_{s}|z_{t}, z_{h(K)})}{p_\theta(z_{s}|z_t)}}_\text{diffusion loss}  - \log p_\theta(z_{h(K)} | z_{h(K)+1}) \Bigg ]
\end{align}
The first and third term can be stochastically and differentially estimated using standard techniques. Following \cite{vdms}, we derive an estimator for the diffusion loss $\mathcal{L}_{(h(K), T)}(z_{h(K)})$. In the case of finite timesteps $t>h(K)$, this loss is: 

\begin{align}
\sum_{t > h(K)} \mathbb{E}_{q(z_{t}|z_{h(K)})} D [q(z_{s}|z_{t}, z_{h(K)}) || p_\theta(z_{s}|z_t) ]
\end{align}
Reparametering $z_t \sim q(z_t | z_{h(K)})$ as $z_t = \alpha_{t|h(K)} z_{h(K)} + \sigma_{t|h(K)} \epsilon$, where $\epsilon \sim \mathcal{N}(0,1)$, and to avoid having to compute all $T - h(K)$ terms when calculating the diffusion loss, we construct an unbiased estimator as: %
\begin{align}
     \frac{T- h(K)}{2} \mathbb{E}_{\epsilon, t \sim \mathcal{U}(h(K), T)}[  D(q(z_{s}|z_{t}, z_{h(K)}) || p_\theta(z_{s}|z_t))]
\end{align}
where $\mathcal{U}(h(K), T)$ is a uniform distribution to sample $h(K) < t \leq T$ from a non-uniform descretization of timesteps using \cite{edm}.

Now, we elaborate on the expression $q(z_s | z_t, z_{h(K)})$ and $p(z_s | z_t)$ for any $h(K) < s < t < T $. 

\subsubsection{\texorpdfstring{$q(z_s | z_t, z_{h(K)})$}{}}
Our posterior at $h(K)$ is $q(z_{h(K)}) = \mathcal{N} (\mu_{h(K)} , \tau_{h(K)}^2)$. For any $h(K) < s < t < T$, we have $q(z_s | z_{h(K)}) = \mathcal{N}(\alpha_{s|h(K)} z_{h(K)}, \tau_{s|h(K)}^2)$ and  $q(z_t|z_s) = \mathcal{N}(\alpha_{t|s} z_s, \sigma_{t|s}^2)$, yielding the posterior :
\begin{align}
    q(z_s | z_t, z_{h(K)}) = \mathcal{N}(\mu_Q(z_t, z_{h(K)}; s,t, h(K)), \sigma_Q^2(s,t, h(K)) I) \\
    \text{where,} \quad \sigma_Q^2(s,t, h(K)) = \sigma_{t|s}^2 \frac{\tau_{s|h(K)}^2}{\sigma_{t|s}^2 + \alpha_{s|h(K)}^2 \tau^2_{s|h(K)}}\\
    \text{and,} \quad \mu_Q(z_t , z_{h(K)}; s,t, h(K)) = \sigma_Q^2 \Bigg (\frac{\alpha_{s|h(K)}}{ \tau^2_{s|h(K)}} z_{h(K)}+ \frac{\alpha_{t|s}}{\sigma_{t|s}^2 }z_t \Bigg ) \\
     = \frac{\alpha_{s|h(K)} \sigma_{t|s}^2}{(\sigma_{t|s}^2 + \alpha^2_{s|h(K)}\tau_{s|h(K)}^2)} z_{h(K)}+ \frac{\alpha_{t|s} \tau_{s|h(K)}^2}{(\sigma_{t|s}^2 + \alpha^2_{s|h(K)}\tau_{s|h(K)}^2)}z_t
\end{align}
\subsubsection{\texorpdfstring{$p(z_s | z_t)$}{}}
The conditional model distributions can be chosen as: 
\begin{align}
   & p_\theta(z_{s}|z_t) = q(z_{s}|z_{t},  z_{h(K)} = \hat z_{\theta, h(K)}(z_t,t))  = \mathcal{N}(z_{s} ; \mu_\theta(z_t, z_{h(K)}; s, t, h(K)), \sigma_Q^2(s,t, h(K)))  \\
   & \text{where,} \quad  \mu_\theta(z_t, z_{h(K)}; s,t, h(K)) =  \frac{\alpha_{s|h(K)} \sigma_{t|s}^2}{(\sigma_{t|s}^2 + \alpha^2_{s|h(K)}\tau_{s|h(K)}^2)} \hat z_{\theta, h(K)}(z_t,t) + \frac{\alpha_{t|s} \tau_{s|h(K)}^2}{(\sigma_{t|s}^2 + \alpha^2_{s|h(K)}\tau_{s|h(K)}^2)}z_t \\
   & \text{and,}  \quad \sigma_Q^2(s,t, h(K)) = \sigma_{t|s}^2 \frac{\sigma_{s|h(K)}^2}{\sigma_{t|s}^2 + \alpha_{s|h(K)}^2 \sigma^2_{s|h(K)}}
\end{align}
where $\hat z_{\theta, h(K)}(z_t, t) = \frac{z_t - \sigma_{t|h(K)} * \epsilon_\theta(z_t,t) }{\alpha_{t|h(K)}}$

\section{Expanded Figure}
\label{app:intuition}
\begin{figure}
\begin{subfigure}{\linewidth}
   \centerline{\includegraphics[width=\linewidth]{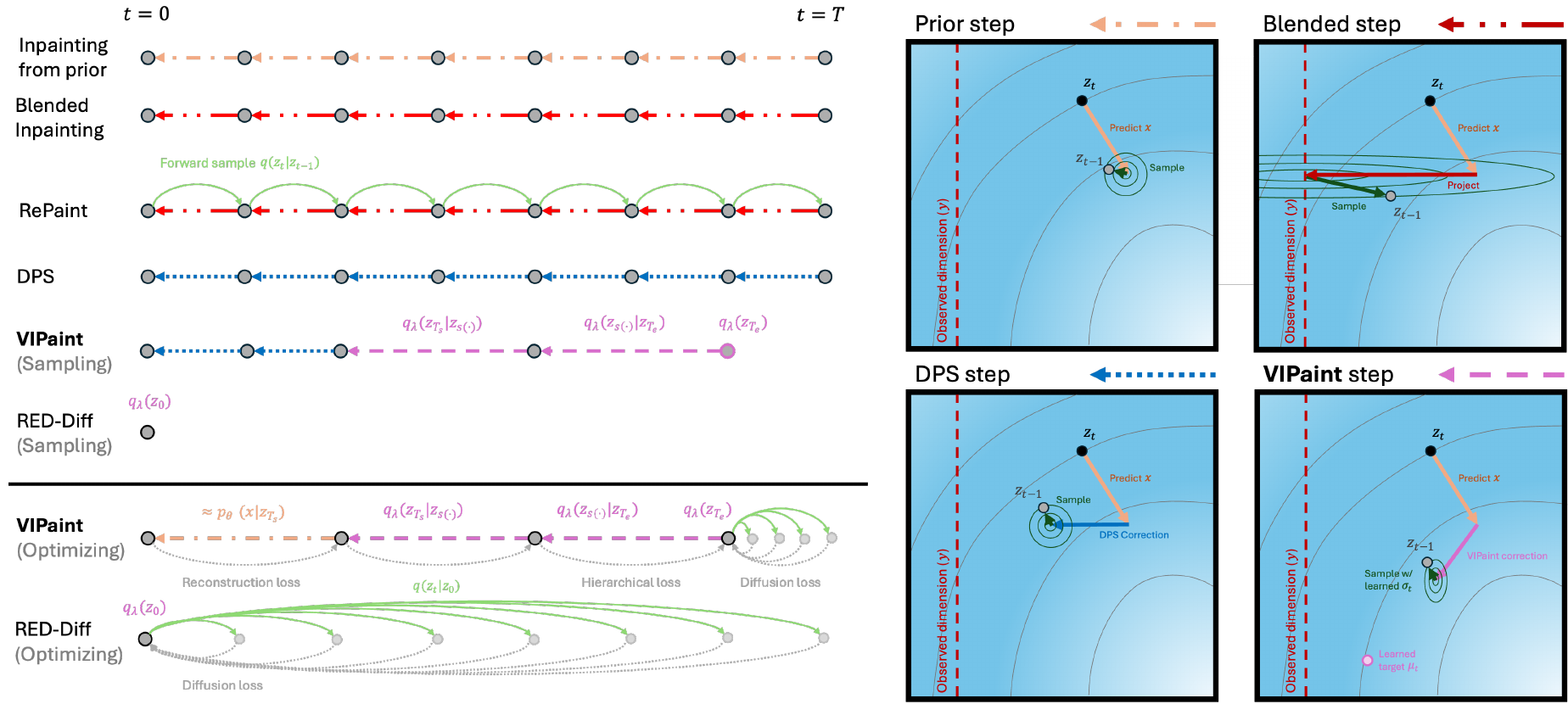}}
 \end{subfigure}
  \caption[]
{\small  Expanded comparison of methods for diffusion model-based inpainting. \textbf{Left:} Timeline illustration of sampling steps with time flowing rightward from $t=0$ (clean images) to $t=T$ (pure noise). \textcolor{orange}{Orange arrows} indicate a single step of ancestral sampling under the generative prior $p_\theta(z_{t-1}|z_{t})$. \textcolor{red}{Red arrows} indicate a single step of the \textit{Blended} approximation of $p_\theta(z_{t-1}|z_{t}, y)$, while \textcolor{blue}{blue arrows} indicate a single step of the \textit{DPS} approximation. \textcolor{green}{Green arrows} indicate steps forward in time according to the diffusion process $q(z_t|z_{<t})$. Methods such as \textit{RePaint} and \textit{CoPaint} alternate between forward and reverse steps. \textcolor{violet}{Purple arrows} indicate sampling from a step in the hierarchical \textbf{VIPaint} posterior $q_\lambda(z_{s(i-1)}|z_{h(i)})$. Both \textbf{VIPaint} and \textit{RED-Diff} (without annealing) involve an initial optimization stage to fit variational parameters per-image. \textcolor{gray}{Gray arrows} indicate the flow of gradient information during this optimization stage. \textcolor{gray}{Gray points} are steps only used during optimization. \textbf{Right:} Illustration of each reverse-time sampling step in 2 dimensions. The horizontal dimension is assumed to be observed at the value marked by the \textcolor{red}{red line}. Each approach begins by computing $p_\theta(z_{t-1}|z_t)$ via a prediction of $x$ using the pre-trained denoising network $\hat{x}_{\theta}(z_t, t)$. \textit{Blended} replaces observed dimensions with $q(z_{t-1}|y)$. \textit{DPS} updates $p_\theta(z_{t-1}|z_t)$ according to a single-step approximation to the likelihood $p_\theta(y|z_{t-1})$. Finally, \textbf{VIPaint}, uses a learned variational distribution $q_\lambda(z_{t-1}|z_t)$, which can be seen as interpolating between the prediction of $x$ and a variational parameter $\mu_t$, coupled with a learned variance.} 
\label{fig:intuition_expanded} 
\end{figure}

We provide an intuitive plot that compares the VIPaint with existing methods that uses pre-trained diffusion models for image inpainting in Fig. \ref{fig:intuition_expanded}. 

\newpage
\section{Experimental Details}
\label{sec:exp-details}

\subsection{VIPaint}
\paragraph{VIPaint-4} Keeping the two endpoints $h(1), h(K)$ the same, we define the hierarchical posterior over timesteps, $[h(K) = 5, 4, 3.5, 2.5, h(1) = 2 ]$ for the EDM noise schedule and $[h(K) = 550, 500, 450, h(1) = 400]$ for the LDM prior.
\paragraph{Initialization} We follow the forward and reverse diffusion process defined by each VE and VP noise schedules to initialize VIPaint's variational parameters. For LDM prior, we use the lower dimensional encoding of $y$. We provide a comprehensive summary in Table \ref{tab:initialization}.
 
\begin{table}[htbp]
\scriptsize
\centering
\caption{Initialization of Variational Parameters for VE and VP Schedules}
\label{tab:initialization}
\begin{tabular}{lcc}
    \toprule
    \textbf{VI Parameters} & \textbf{VP Schedule} & \textbf{VE Schedule} \\
    \midrule
    \begin{tabular}[c]{@{}l@{}} $\mu_{h(K)} = \alpha_{h(K)} y + a_1 \sigma_{h(K)} \epsilon$ \\ (Scale factor to retain information from $y$.) \end{tabular} 
    &  $a_1=0.8$ & $a_1 = 0.01$ \\
   \begin{tabular}[c]{@{}l@{}} $\mu_{h(i)} = \alpha_{h(i)} y + a_2 \sigma_{h(i)} \epsilon$  \\ (Noise adding process is still quite high \\ for VE schedules.) \end{tabular}&  $a_2 = 1$ &  $a_2 = 0.01$ \\
   \begin{tabular}[c]{@{}l@{}} $\tau_{h(K)} = \sigma_{h(K)}$ \\ (From the forward diffusion process. )\end{tabular} & --
     &  -- \\
    \begin{tabular}[c]{@{}l@{}} $\tau_{h(i) \mid h(i+1)}$ \\ (From the reverse diffusion process.) \end{tabular}  & \begin{tabular}[c]{@{}l@{}} Eq. \ref{eq:VP_std} with scaling factor $a_3/\eta$ \\   $a_3 = 0.7$ \end{tabular}  & Eq. \ref{eq:VE_std}\\
    \begin{tabular}[c]{@{}l@{}} $\gamma_{h(i)} \forall i \in [1,K]$  \\ (Weights samples from prior to construct\\ plausible and close to real looking samples.) \end{tabular} & $0.98$ (ImageNet256), $0.88$ (LSUN)  & $0.5$ \\
    \bottomrule
\end{tabular}
\end{table}
 
\paragraph{Optimization}
We fit three sets of variational parameters at every $i$-th critical time in our hierarchy: means, $\mu_{h(i)}$, variances $\tau^2_{h(i)}$ and weights $\gamma_{h(i)}$. Instead of optimizing $\tau$ and $\gamma$ directly, we optimize the real valued $\tilde \tau =  \log \tau^2$, and $\tilde \gamma = \log(\frac{\gamma}{1- \gamma})$. We optimize this set of variational parameters $\lambda = \{\mu, \tilde \gamma, \tilde \tau\} $ using Adam with an initial learning rate of $\{0.1, 0.1, 0.01\}$ respectively and decreasing the learning rate by a factor of $0.99$ every $10$ iterations. We find this setting to be robust across all prior diffusion models and datasets in our work. 

\begin{figure} 
\centerline{
\includegraphics[width=0.7\linewidth]{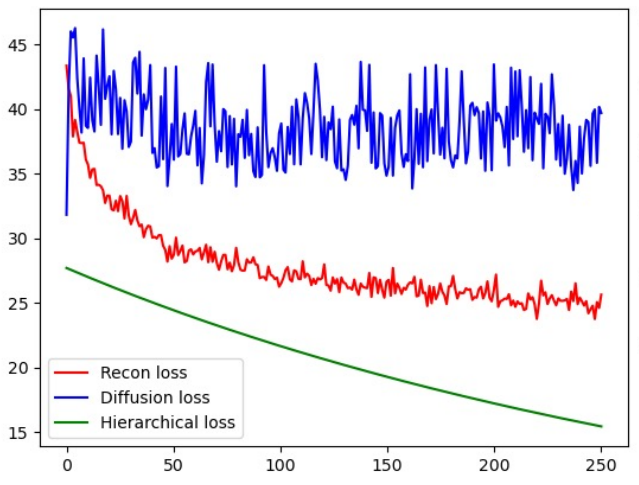}}
  \caption[]
{\small Loss plot over $250$ optimization steps for fitting the hierarchical posterior of VIPaint-2.
} 
\label{fig:loss_plot}
\end{figure}

During pre-training, most diffusion models parameterize the mean prediction at every diffusion time step $t$ and fix variances, however some previous work \citep{ddpm++,dhariwal2021diffusion} has found that (with appropriate training “tricks”) learning variances improves performance. Some previous works like ReSample tunes this as a hyperparameter. We instead learn this in our work, and we adjust learning rates to avoid local optima in this process. We optimize the parameters in VIPaint with $K=2$ for $50$ iterations (we show a loss curve in Fig \ref{fig:loss_plot}); VIPaint with $K=4$ is optimized for $100$ steps in the case of LSUN Churches, $150$ steps for the ImageNet64 dataset and $250$ steps for the ImageNet256 dataset. 

In the case of Stable Diffusion 3.5, a text-conditioned model, we optimized VIPaint with $K=2$ hierarchical levels for $50$ iterations. The corresponding hierarchical bounds were set to $h(1) = 250$ and $h(2) = 500$. We used a learning rate of $0.2$ and targeted a resolution of $1024 \times 1024$. The training objective from Eq.~\eqref{eq:VIPaintobj} combines pixel-space reconstruction loss ($L_1$, weight $0.05$), latent-space reconstruction loss ($L_2$, weight $6.0$), diffusion loss (weight $0.5$), and hierarchical loss (weight $0.5$). We empirically found that this combination, together with a DPS guidance scale of $650.0$, provides the best trade-off between reconstruction fidelity and contextual consistency.

\paragraph{Sampling} Post training, we take $400$ iterative refinement steps from $h(1) = 400$ in the LDM variance preserving schedule to sample inpaintings using a scale factor of $2$, similar to the DPS algorithm using perceptual loss. On the other hand, for the EDM prior, we take $700$ refinement steps to produce inpaintings after $h(1) = 2$, with scale $5$ (similar to DPS tuned for EDM in our work). This scale hyperparameter is tuned over the values [0.1, 0.5, 1, 2, 5, 10] on a validation set of 20 images. During the sampling phase,  we use the classifier-free guidance rule with scale $=3$ for the ImageNet256 latent diffusion prior.
\begin{figure}
\centering   
\includegraphics[width=\linewidth]{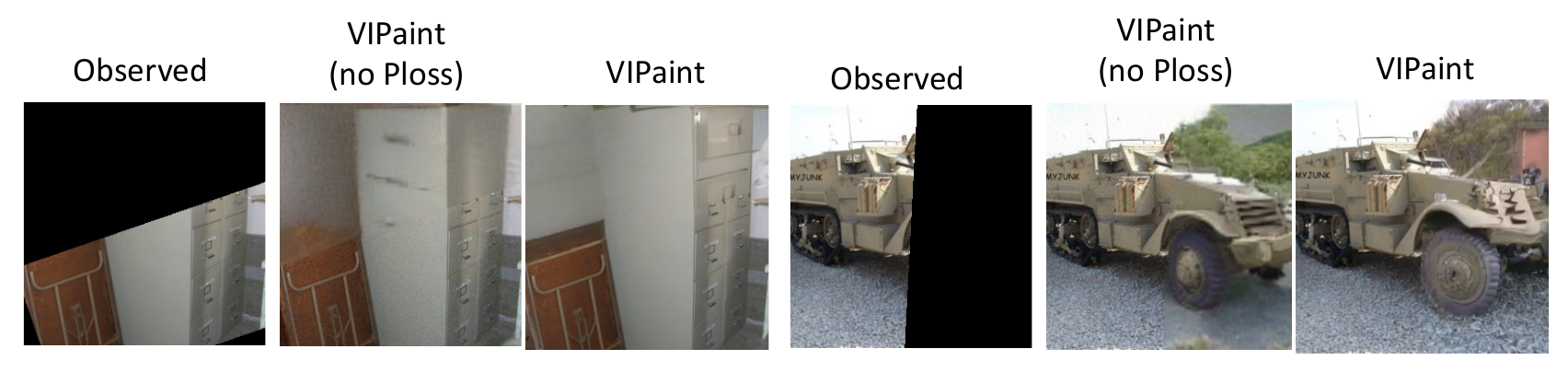}
    \caption{An ablation showing the effect of addition the perceptual loss (PLoss) in the reconstruction term for the task of image inpainting using latent diffusion priors. We see that even though VIPaint can inpaint the image semantically without the Perceptual loss, this loss becomes important to produce sharper reconstructions.}
\label{fig:ploss_ablation}
\end{figure}
\begin{figure} 
\centerline{
\includegraphics[width=\linewidth]{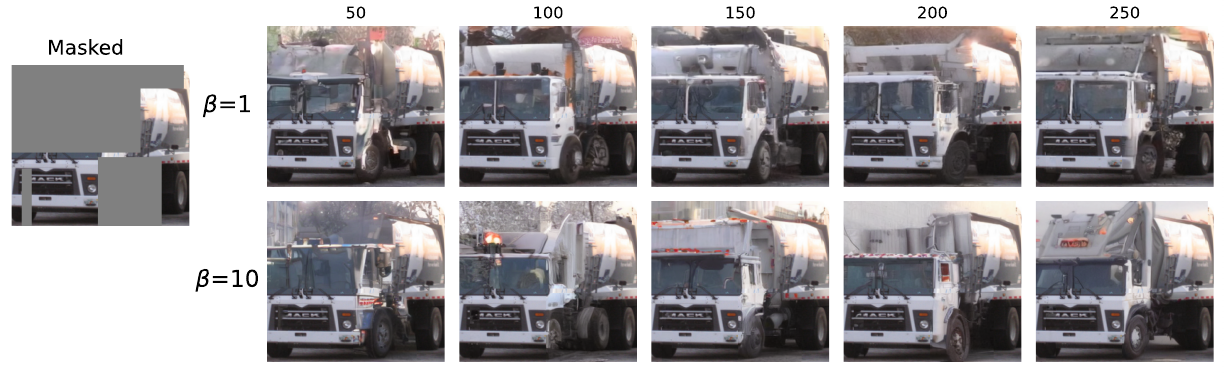}}
  \caption[]
{\small We show the effect of the hyperparameter $\beta$ with VIPaint with respect to optimization iterations. %
With $\beta=10$, VIPaint captures more variations under the diffusion prior instead of "setting" to one kind of completion with $\beta=1$. 
} 
\label{fig:ablations_beta}
\end{figure}

\paragraph{Reconstruction loss} We assume $p(y|z_{h(1)})$ as a Laplacian distribution, where the mean is given by $y$ and a scale parameter, which is computed over $100$ images per dataset as a standard deviation over all pixel dimensions. For the $256$ pixel datasets, this is $0.56$, and for ImageNet64 it is $0.05$. In addition to this, we add the perceptual loss for LDM priors, computing them via feeding the pre-trained Inception network with masked images and masked reconstructions. See Fig. \ref{fig:ploss_ablation} for the benefits of using the perceptual loss with VIPaint. 

For VIPaint with $K=4$, we upweight the KL loss terms in VIPaint's objective with a weight $\beta=50$ for pixel-based EDM prior and $\beta=10$ for LDM prior. We show the effect of the different $\beta$ values in Fig. \ref{fig:ablations_beta}. Generally speaking, higher values of $\beta$ explores the diffusion latent space more and lower values weighs the likelihood term relatively more and converges faster to a solution. 

\paragraph{Descretization of timesteps for prior diffusion loss} Lastly, we directly adapt the descretization technique from EDM \cite{edm} to compute the diffusion loss. We use $\rho = 7$ across all models and datasets as used by \cite{edm}.
\subsection{Baseline Details}
\label{sec:appendix-baselines}
Across all the baselines applicable to the latent diffusion models for the ImageNet256 dataset, we use the classifier-free guidance with a scale of $3$ \citep{Rombach_2022_CVPR}.

\paragraph{Blended.} We run blended for $1000$ discretization steps using the EDM and LDM prior. 

\paragraph{RePaint.} RePaint uses a descritization of $256$ steps along with the standard jump length = 10, and number of times to perform this jump operation also set to 10, following standard practice \cite{Lugmayr2022RePaintIU}. 

\paragraph{DPS.} Similar to blended, we take $1000$ denoising steps for DPS and set scale = $5$ for the edm-based diffusion model, while take $500$ steps and keep scale as $0.5$ for the Latent Diffusion prior (similar to the original work in \cite{chung2023diffusion}). When using the perceptual loss for the latent diffusion prior, we increase the scale to 2.

\paragraph{PSLD.} This is an inference technique only for the Latent Diffusion prior. Similar to DPS, we take $500$ steps and keep scaling hyperparameters set to $0.2$ as opposed to choosing $0.1$ in the original paper \cite{rout2023solving}. We observe artifacts in the inpainted image if we increase the scale further as also observed by \cite{chung2024prompttuning}.  

\paragraph{CoPaint.} We directly adapt the author-provided implementation of CoPaint and CoPaint-TT \cite{pmlr-v202-zhang23q} to use the EDM prior. Apart from the diffusion schedule and network architecture (taken from EDM) all other hyperparameters are preserved from the base CoPaint implementation. 

\paragraph{RED-Diff.} As with CoPaint, We directly adapt the author-provided implementation of RED-Diff \cite{mardani2024variational} 
 and Red-Diff (Var) to use the EDM prior. In this case we increased the prior regularization weight from 0.25 to 50, which we found gave improved performance and more closely matches our VIPaint settings.

\paragraph{ReSample.} As with other baselines, we directly adapt the author-provided implementation of ReSample \cite{resample} for the LDM prior. Because the original code takes larger optimization steps, resulting in high sampling time, we decrease the number of optimization steps to $50$, such that the wall-clock run-time of this method matches the other baselines. 

\paragraph{SD-Inpainting.} We use the official \texttt{StableDiffusionInpaintPipeline} from the Diffusers library with default settings. The model \texttt{stabilityai/stable-diffusion-2-inpainting}~\citep{Rombach_2022_CVPR} is used with $50$ inference steps to match the optimization steps in our method.  

\paragraph{ControlNet.} We use the \texttt{StableDiffusionControlNetInpaintPipeline} from the Diffusers library with default settings, based on the model proposed by~\citep{zhang2023adding}. We run $50$ inference steps for consistency with our method.  

\section{Inference Time.} 
\label{app:time-analysis}

\textbf{Time Complexity.} The time taken by VIPaint-$K$ to optimize a Markov posterior with $K$ keypoints scales primarily with the number of denoising network calls. Each optimization step (assuming $K \ll T$) involves $O(K)$ calls to sample $z_{h(1)} \sim q_\lambda(z_{h(1):h(K)})$, and one to compute the diffusion prior loss, resulting in $O(K)$ function calls per step. Thus, for $I$ optimization steps, the overall complexity is $O(KI)$. For example, our VIPaint-2 is optimized over $50$ iterations, it requires only $50*(2+1)=150$ denoising network calls to infer global image semantics. %
Once fit, sampling requires an additional $h(1)$ denoising network calls per sample instead of $T$ network calls as in traditional sampling methods.

In tables \ref{tab:runtime_appendix} and \ref{tab:runtime2}, we report the time taken for each inference method to produce $10$ inpaintings for 1 test image. VIPaint with $K=2$ is comparable to the baseline methods in terms of wall clock time and the number of functional evaluations ($E$) of the denoising network. Red-Diff, Blended and RePaint baseline methods do not take gradient of the noise prediction network, whereas all other methods require gradients. In terms of time and number of function calls, we can see that VIPaint-2 takes comparable time and number of function calls as other baselines, but performs far better (Table \ref{tab:performance_metrics_pixel_appendix}). 

Overall, gradient based methods like DPS take longer with an LDM prior because of the use of a decoder per gradient step. PSLD additionally utilizes the encoder and hence, takes longer than DPS. 

\begin{table}[t]
\caption{Runtime Comparison For Inference Methods.}
\vspace*{-6pt}
\scriptsize
\centering
\begin{tabular*}{\linewidth}{@{\extracolsep{\fill}} lc@{\hskip 1pt}c@{\hskip 1pt}c@{\hskip 5pt}c@{\hskip 1pt}c@{\hskip 5pt}c@{\hskip 1pt}c@{\hskip 5pt}c@{\hskip 1pt}c@{\hskip 5pt}c@{\hskip 1pt}c@{\hskip 5pt}c@{\hskip 1pt}c@{\hskip 5pt}c@{\hskip 1pt}c@{\hskip 5pt}c@{\hskip 1pt}c@{\hskip 5pt}c}
    \toprule
    Dataset & \textbf{Blended} & \textbf{DPS} & \textbf{VIPaint} & \textbf{\emph{Sample}} \\
    \midrule
   ImageNet64 &  (1.13, 1000) & (2.55, 1000) &  (1.5, 150) & (1.8, 700)  \\
    \midrule
    ImageNet256 &  (4, 1000) & (10, 500) & (2, 150)   & (8, 400) &  \\ 
    LSUN &  (1.3, 1000) & (5.1, 500)  & (2.1, 150)  &(4.3, 400)  
    \\ \bottomrule
\end{tabular*}
\caption*{The (\textit{time in minutes}, \textit{neural function evaluations}) are reported for EDM (top) and LDM (bottom) priors. 
For VIPaint, optimization (``VIPaint") and sampling are separated, since optimized posterior can be reused. 
RedDiff matches Blended, while RePaint (2.8 mins) and CoPaint (2.6 mins) are slightly slower than DPS.}
\label{tab:runtime_appendix}
\end{table}

\begin{table}[b]
\scriptsize
\centering
\begin{tabular*}{\linewidth}{@{\extracolsep{\fill}} lc@{\hskip 1pt}c@{\hskip 1pt}c@{\hskip 5pt}c@{\hskip 1pt}c@{\hskip 5pt}c@{\hskip 1pt}c@{\hskip 5pt}c@{\hskip 1pt}c@{\hskip 5pt}c@{\hskip 1pt}c@{\hskip 5pt}c@{\hskip 1pt}c@{\hskip 5pt}c@{\hskip 1pt}c@{\hskip 5pt}c@{\hskip 1pt}c@{\hskip 5pt}c}
    \toprule
    Dataset & \textbf{VIPaint-2 (opt.)} & \textbf{VIPaint-2 (sample)} & \textbf{VIPaint-4 (opt.)} & \textbf{VIPaint-4 (sample)} \\
    \midrule
   ImageNet64 &  (1.5, 150) & (1.8, 700) &  (10.0, 900) & (1.8, 700)  \\
    \midrule
    ImageNet256 & (2, 150)   & (8, 400) &  (10.0, 1250) & (8.0, 400) \\ 
    LSUN   & (2.1, 150)  &(4.3, 400)  &  (5.5, 750) & (4.3, 400)  
\end{tabular*}
\vspace{1mm}
\caption{Table comparing (\textit{time in minutes}, \textit{neural function evaluations}) across inference methods using the EDM Prior (top) and LDM Prior (bottom). For VIPaint, we separate the optimization (opt.) and sampling steps, as the optimized posterior approximation can be reused across samples.}
\label{tab:runtime2}
\end{table}

\begin{table}[t]
\caption{Quantitative ImageNet64 Inpainting Results.}
\vspace*{-8pt}
\small
\centering
\resizebox{\linewidth}{!}{%
\begin{tabular}{@{}lllllllll@{}}
\toprule
           &  \multicolumn{3}{l}{Rotated Window}            & & \multicolumn{3}{l}{Random Mask}                \\ \cmidrule{2-4} \cmidrule{6-8} 
Method     & PSNR$\uparrow$          & SSIM$\uparrow$          & LPIPS$\downarrow$         & & PSNR$\uparrow$           & SSIM$\uparrow$          & LPIPS$\downarrow$         \\ \midrule
VIPaint-2  & \underline{9.24}     & \textbf{0.56}       & \textbf{0.30}         & & \textbf{13.33} & \textbf{0.62}      & \textbf{0.23} \\
CoPaint-TT & 8.51                 & 0.51                & \underline{0.32}      & & \underline{12.51}          & 0.58               & \underline{0.25}          \\
CoPaint    & 8.47                 & 0.50                & 0.35                  & & 12.12          & 0.56               & 0.28          \\
RePaint    & 8.82                 & \textbf{0.56}       & \underline{0.32}      & & 12.05          & \underline{0.59}   & 0.26          \\
DPS        & 8.15                 & \underline{0.53}    & \underline{0.32}      & & 11.45          & 0.56               & 0.29          \\
Blended    & 7.68                 & 0.52                & 0.34                  & & 11.47          & 0.57               & 0.28          \\
RedDiff    & 8.56                 & 0.45                & 0.46                  & & 11.89          & 0.51               & 0.41          \\
RedDiff-V  & \textbf{9.27}        & \underline{0.53}    & 0.41                  & & 8.35           & 0.16               & 0.67          \\ \bottomrule
\end{tabular}%
}
\caption*{
Using the pixel-based EDM prior for all methods, PSNR, SSIM, and LPIPS are averaged over 1000 inpaintings. VIPaint shows the best performance (\textbf{bold}), and the second best is \underline{underlined}.}
\vspace*{-6pt}
\label{tab:performance_metrics_pixel_appendix}
\end{table}

\section{Computational resources} All experiments were conducted on a system with 4 Nvidia A6000 GPUs.

\newpage
\section{Additional Plots}
\label{app:addn_plots}

\subsection{Analysis of Imagenet Results}
\begin{figure}
    \centering
    \includegraphics[width=\linewidth]{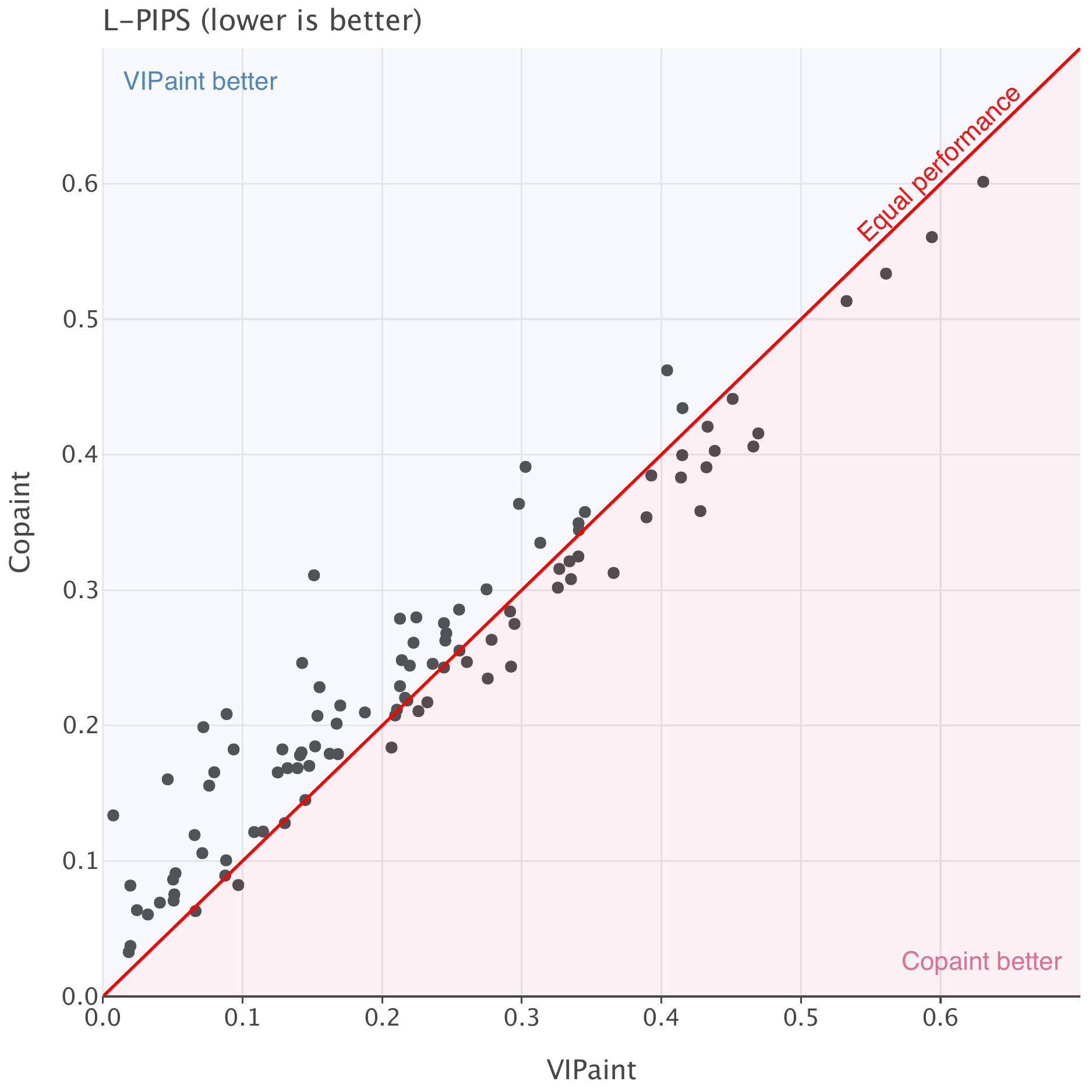}
    \caption{Paired comparison of LPIPS scores for VIPaint-2 and CoPaint with time-travel (CoPaint-TT) on the Imagenet64 ``Random Mask" inpainting task (expanding on the experiment shown in table 1. Each point shows the mean LPIPS score across 10 sampled completions of the masked image, with the x and y coordinates showing the VIPaint and CoPaint-TT scores respectively. Additionally, we validated that VIPaint improves on CoPaint-TT using a one-sided paired t-test on the mean LPIPS scores of each method. We found that the improvement was statistically significant with a p-value of \textbf{4.133e-05}. As the normality assumption of the t-test may not hold, we also verfied the results using a nonparametric Wilcoxon signed ranked test, which indicated a statistically significant improvement with a p-value of \textbf{0.000114}}
    \label{fig:lpips-significance}
\end{figure}

Fig. \ref{fig:lpips-significance} shows details of the comparison between VIPaint and CoPaint with time-travel over 100 randomly selected images from the Imagenet-64 task.

\subsection{Small-Mask Image Inpainting for LSUN, ImageNet256}
\begin{figure}
    \centering    \includegraphics[width=0.8\linewidth]{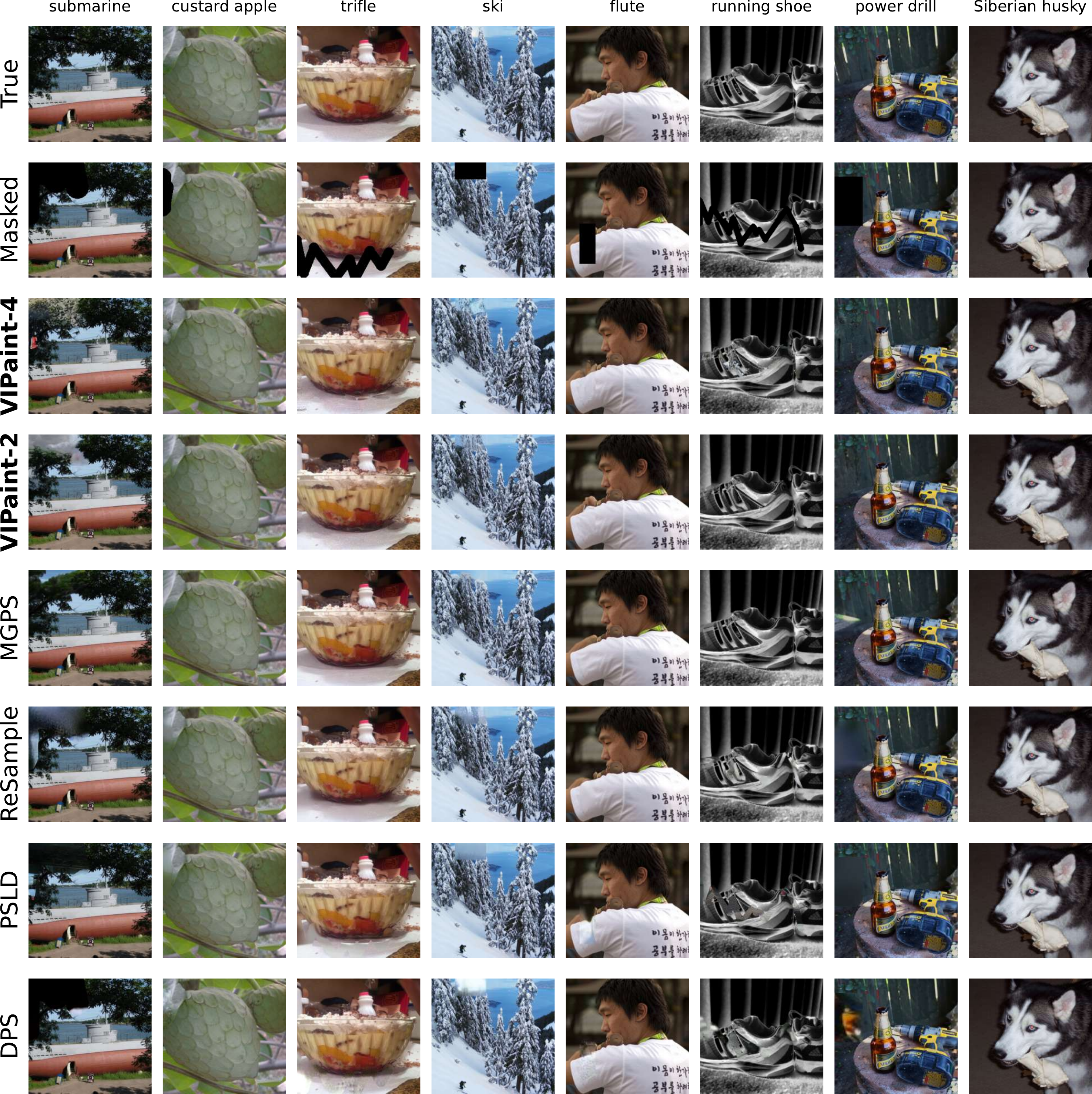}
    \caption{Qualitative results on the performance across methods for small masking ratios for ImageNet256 dataset using LDM prior. All methods seem to perform reasonably well in this regime.}
    \label{fig:small_masks_im}
\end{figure}

\begin{figure}
    \centering    \includegraphics[width=0.8\linewidth]{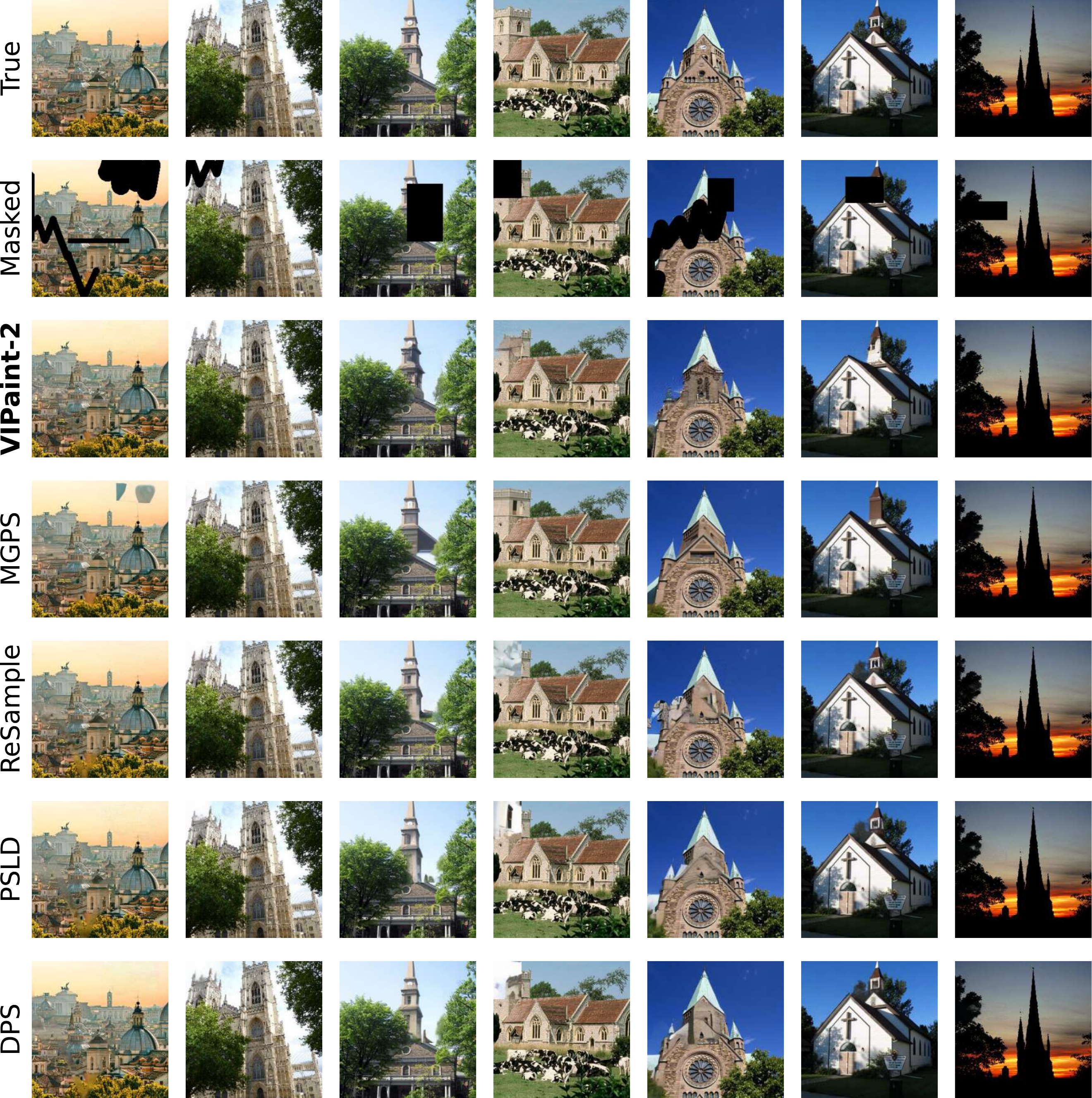}
    \caption{Qualitative results on the performance across methods for small masking ratios for LSUN dataset using LDM prior. All methods seem to perform reasonably well in this regime.}
    \label{fig:small_masks_lsun}
\end{figure}

We show some qualitative figures for small masking ratios (upto $20\%$ of the image is corrupted) in Fig. \ref{fig:small_masks_im} for ImageNet-256 and \ref{fig:small_masks_lsun} for LSUN dataset.

\subsection{VIPaint captures multi-modal posterior}
\begin{figure}
\begin{subfigure}[t]{\linewidth}
\centerline{\includegraphics[width=\linewidth]{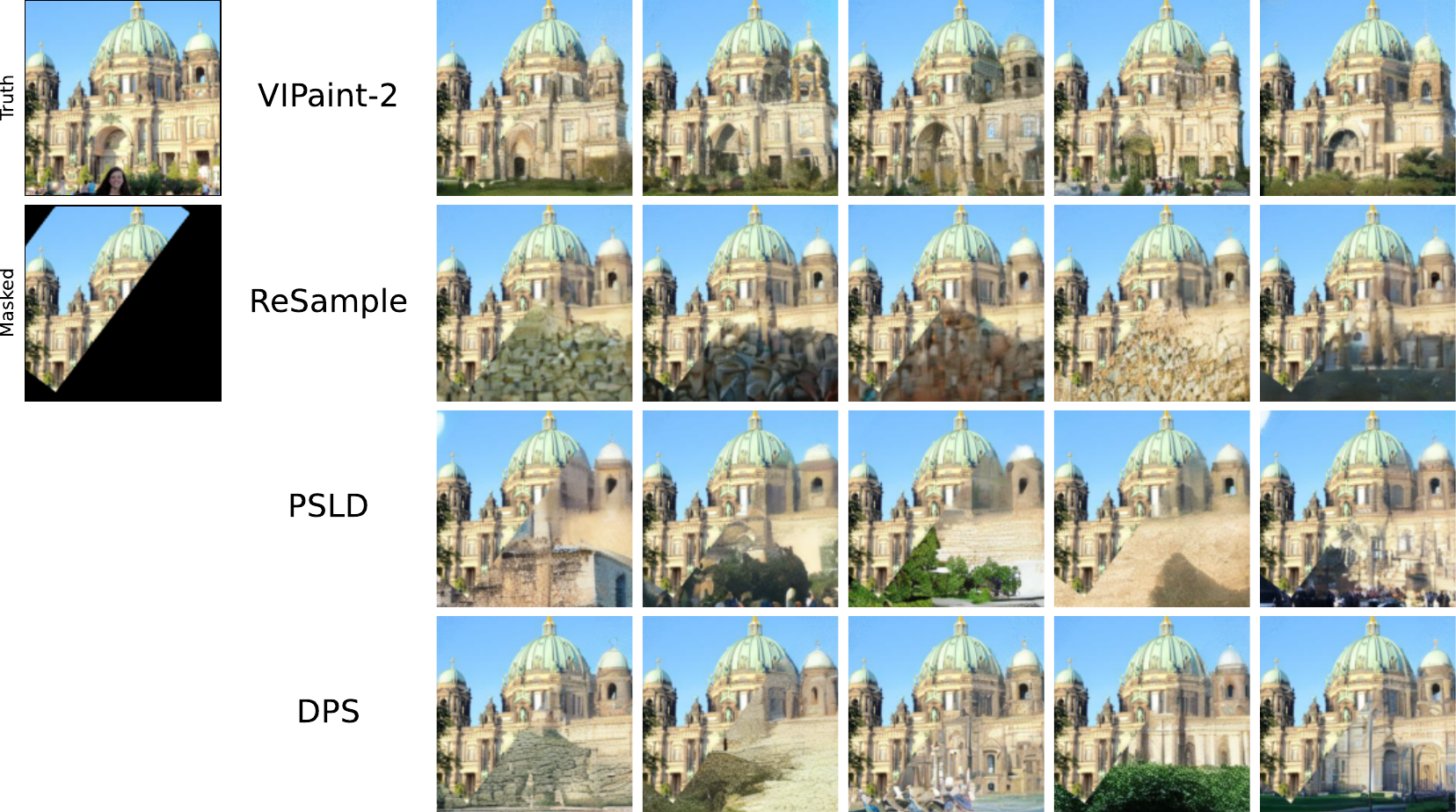}}
 \end{subfigure}
  \caption[]
{LSUN diversity results. Examples of diverse generation using VIPaint and baseline methods on LSUN using the same input and different initial noise. 
} 
\label{fig:lsun256-variations}
\end{figure}

\begin{figure*}[t!]
 \begin{subfigure}[t]{0.9\linewidth}
\centerline{\includegraphics[width=0.9\linewidth]{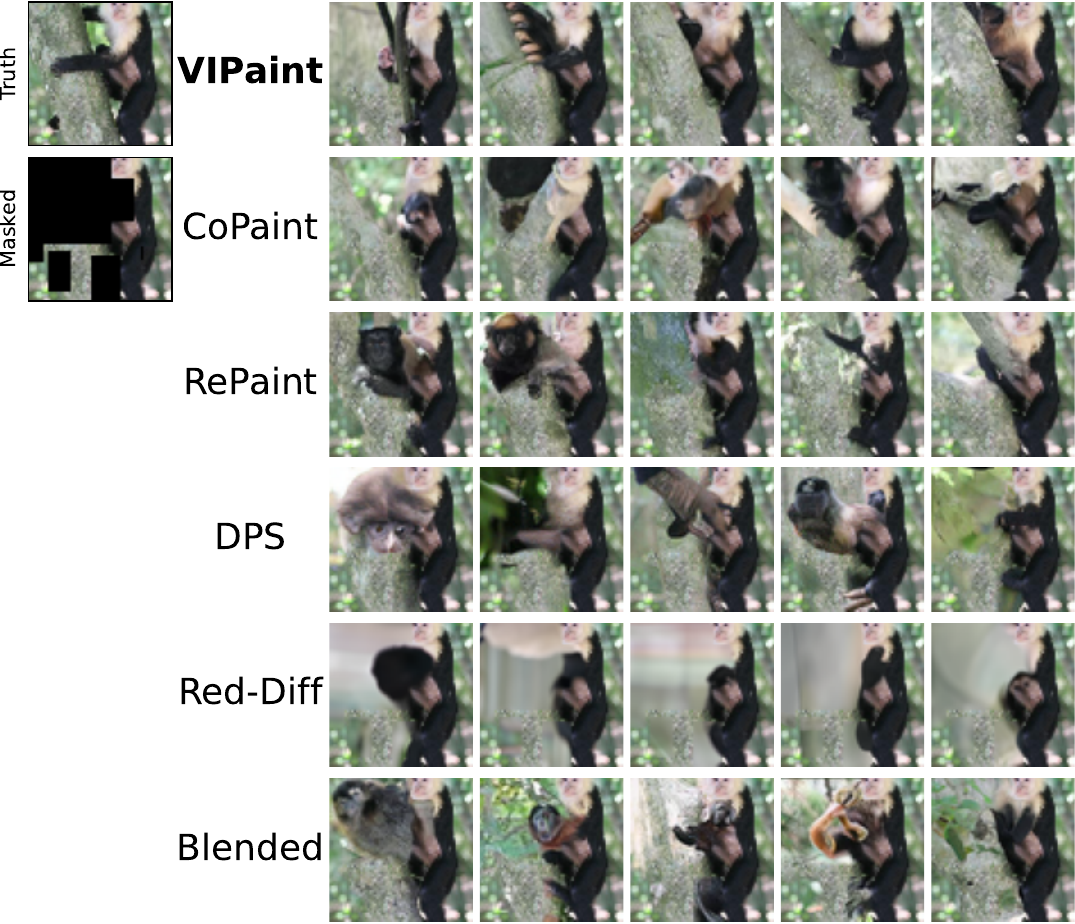}}
 \end{subfigure}
  \caption[]
{ImageNet64 diversity results with the same class condition but different initial noise.
} 
\label{fig:variations_last}
\end{figure*}

In addition to producing valid inpaintings, we show multiple samples per test image for all datasets we consider in Fig. \ref{fig:lsun256-variations}
-\ref{fig:variations_last}.  

\subsection{More qualitative results}
\begin{figure}[t] 
  \centering
  \includegraphics[width=\linewidth]{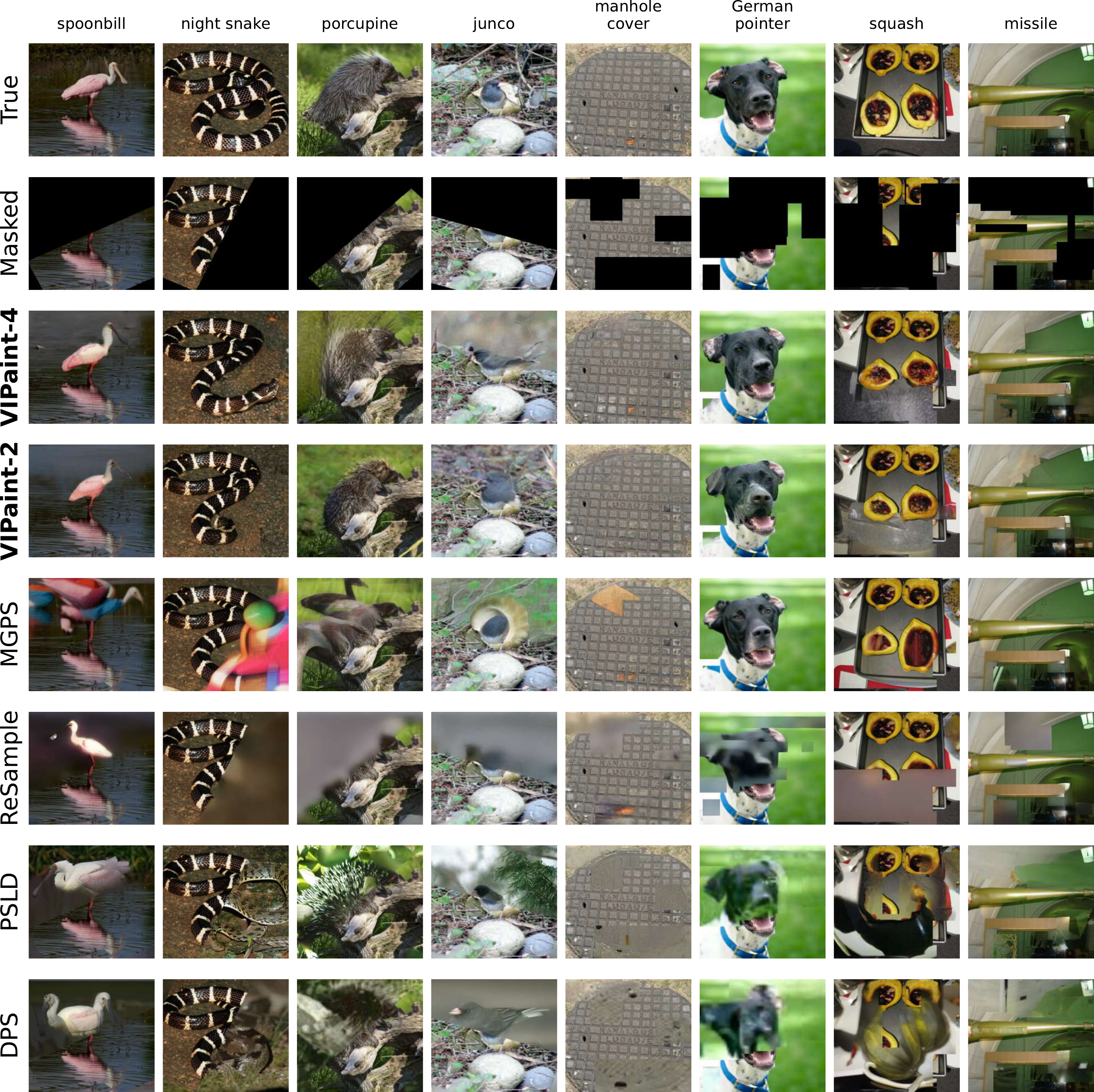}
  \caption{Image completion results on Imagenet256 using the LDM prior for Rotated Window and Random Masking schemes shown in the second row. We show an inpainting from each method in the following four rows. DPS, PSLD, and ReSample show blurry inpaintings of widely varying quality.  In contrast, VIPaint interprets the global semantics in the observed image and produces \emph{very} realistic images. Please find more qualitative plots for LSUN-church in the Appendix Fig.~\ref{fig:consistency-lsun}.
} 
\label{fig:consistency-imagenet256}
\end{figure}

\begin{figure}
    \centering
\includegraphics[width=\linewidth]{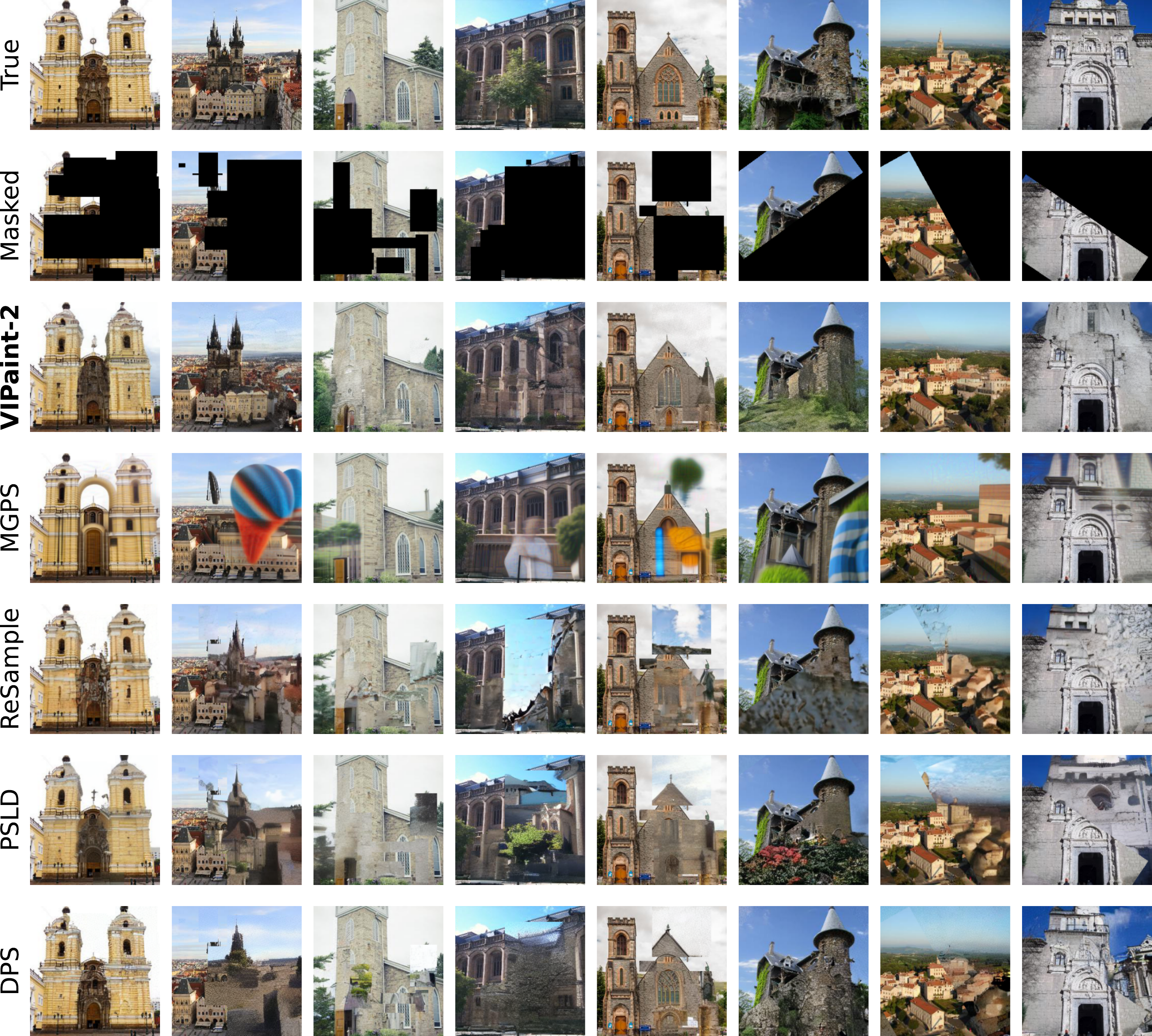}
    \caption{Qualitative results for LSUN-church dataset using LDM prior for the tasks of image inpainting with large masks. We see that VIPaint-2 can inpaint the images consistently and without any artifacts at the mask borders.}
    \label{fig:consistency-lsun}
\end{figure}

\begin{figure}[t] 
  \centering
  \includegraphics[width=.9\linewidth]{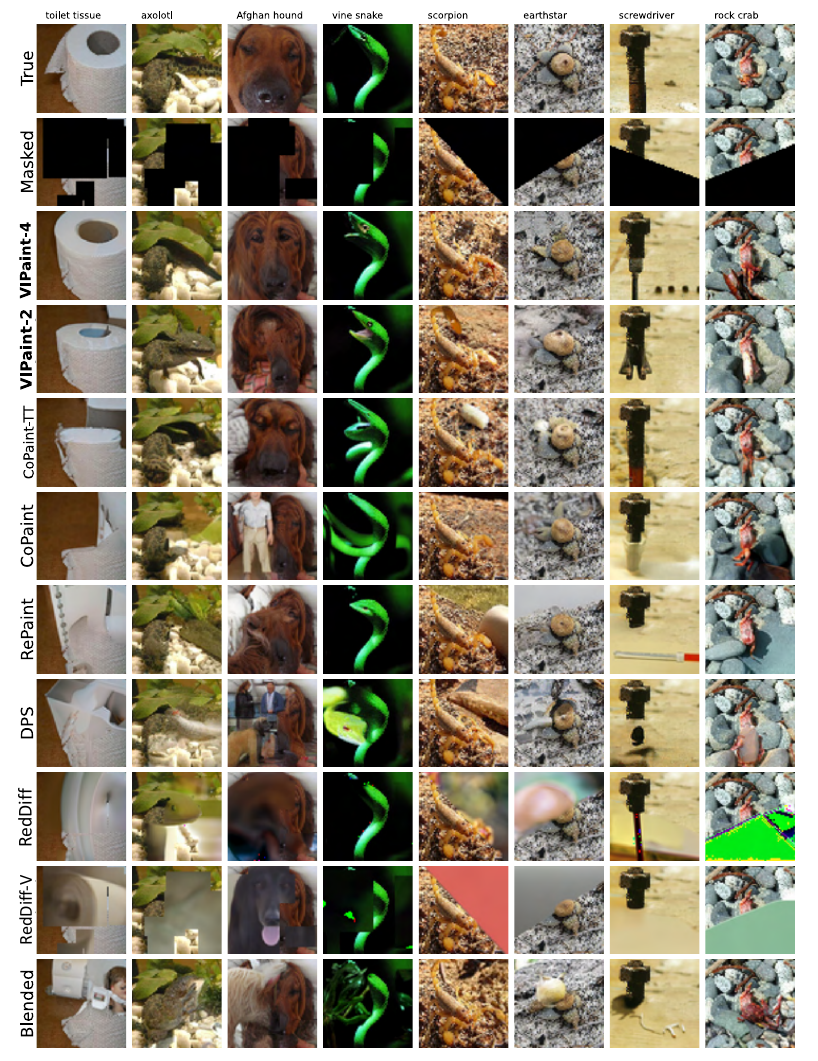}
  \vspace*{-2pt}
  \caption{Image completion results on ImageNet64 using a conditional pixel-based EDM prior for image inpainting (Random Masking and Rotated Window schemes) shown in the second row. We show an inpainting from each method in the following rows.
Even though the prior diffusion model for ImageNet is conditioned on class labels, inpaintings for baseline methods are inconsistent with the observed image. RePaint and CoPaint is typically more accurate than other baselines, but still produce inconsistent samples unless masks are small. In contrast, VIPaint interprets the global semantics in the observed image while enforcing consistency with the few observed pixels. 
} 
\label{fig:consistency-imagenet64}
\end{figure}

We provide more test examples for large mask inpainting in Fig. \ref{fig:consistency-imagenet256}, \ref{fig:consistency-lsun}, \ref{fig:consistency-imagenet64}. 

